\documentclass{article}

\usepackage{microtype}
\usepackage{graphicx}
\usepackage{booktabs} 
\usepackage{xcolor}
\usepackage{caption}
\usepackage{subcaption}

\usepackage[backref=page]{hyperref}

\usepackage[accepted]{icml2024}

\usepackage{amsmath}
\usepackage{amssymb}
\usepackage{mathtools}
\usepackage{amsthm}

\usepackage{pifont}

\usepackage[capitalize,noabbrev]{cleveref}

\theoremstyle{plain}

\theoremstyle{definition}

\theoremstyle{remark}

\usepackage[textsize=tiny]{todonotes}

\icmltitlerunning{
Hybrid Neural Representations for Spherical Data
}

\begin{document}

\twocolumn[
\icmltitle{Hybrid Neural Representations for Spherical Data}



\icmlsetsymbol{equal}{*}

\begin{icmlauthorlist}
\icmlauthor{Hyomin Kim}{yyy}
\icmlauthor{Yunhui Jang}{yyy}
\icmlauthor{Jaeho Lee}{yyy}
\icmlauthor{Sungsoo Ahn}{yyy}
\end{icmlauthorlist}

\icmlaffiliation{yyy}{Pohang University of Science and Technology}

\icmlcorrespondingauthor{Hyomin Kim}{hyomin126@postech.ac.kr}
\icmlcorrespondingauthor{Yunhui Jang}{uni5510@postech.ac.kr}
\icmlcorrespondingauthor{Jaeho Lee}{jaeho.lee@postech.ac.kr}
\icmlcorrespondingauthor{Sungsoo Ahn}{sungsoo.ahn@postech.ac.kr}

\icmlkeywords{Machine Learning, ICML}

\vskip 0.3in
]



\printAffiliations{} 

\begin{abstract}
In this paper, we study hybrid neural representations for spherical data, a domain of increasing relevance in scientific research. In particular, our work focuses on weather and climate data as well as comic microwave background (CMB) data. Although previous studies have delved into coordinate-based neural representations for spherical signals, they often fail to capture the intricate details of highly nonlinear signals. To address this limitation, we introduce a novel approach named Hybrid Neural Representations for Spherical data (HNeR-S). Our main idea is to use spherical feature-grids to obtain positional features which are combined with a multilayer perception to predict the target signal. We consider feature-grids with equirectangular and hierarchical equal area isolatitude pixelization structures that align with weather data and CMB data, respectively. We extensively verify the effectiveness of our HNeR-S for regression, super-resolution, temporal interpolation, and compression tasks. 
\end{abstract}

\section{Introduction}
\begin{figure*}[t]
\centering
    \includegraphics[width=\textwidth]{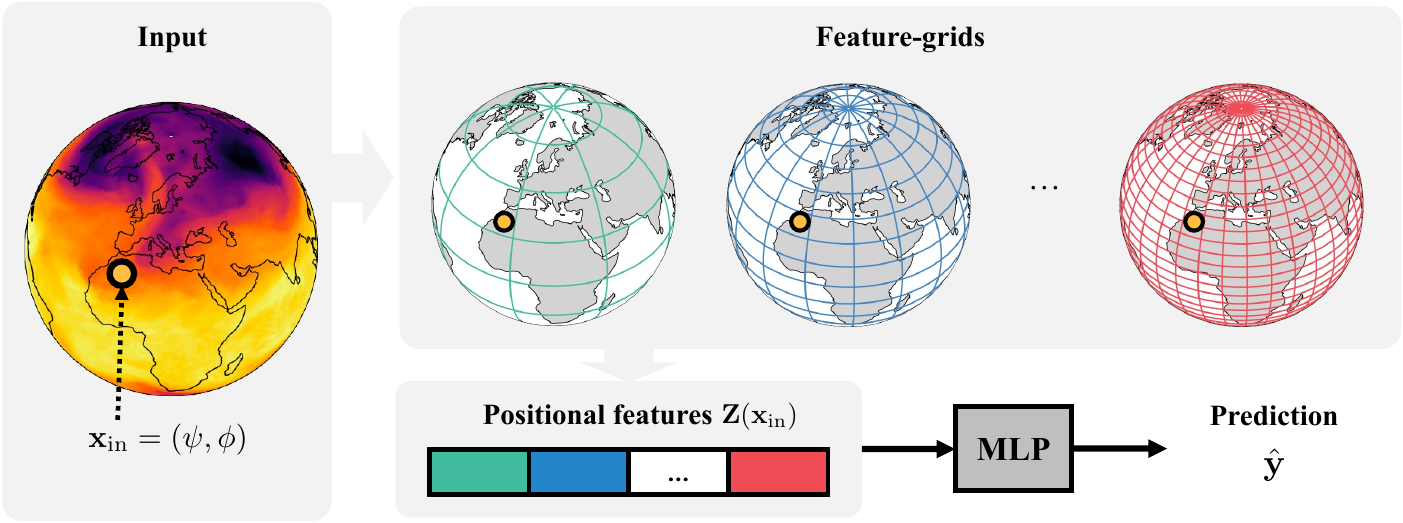}
    \caption{\textbf{Overview of hybrid neural representation for spherical data (HNeR-S).} HNeR-S considers an input point as the spherical coordinate $\mathbf{x}_{\text{in}}=(\psi, \phi)$, a pair of latitude $\psi \in [-\frac{1}{2}\pi, \frac{1}{2}\pi]$ and longitude $\phi \in [0, 2\pi)$. Then the model interpolates the neighborhood feature-grid parameters and constructs the positional features $\mathbf{Z}(\mathbf{x}_{\text{in}})$. The MLP predicts the target signal values from the positional features.}
    \label{fig:overview}
    \vspace{-0.1in}
\end{figure*}

Coordinate-based neural representations (CNRs) form a family of techniques to parameterize the target signal using a neural network that maps input coordinates to signal values \citep{nfvisual}. This results in a continuous function that seamlessly interpolates discretized coordinates of the target signal. As a case in point, when applied to image representation, CNRs trained to fit RGB values on 2D-pixel coordinates can also predict the continuum of coordinates between pixels. This capability has been proven to be useful in representing, interpolating, and increasing the signal resolutions of various modalities, including images \citep{chen2021learning}, signed distance functions \citep{park2019deepsdf}, and radiance fields \citep{mildenhall2021nerf} as special cases.

However, CNRs have been less investigated for spherical signals despite the important scientific applications, e.g., global weather and climate modeling \citep{hersbach2018era5} or the analysis of cosmic microwave background (CMB) radiation~\citep{de2000flat}. This is noteworthy given the necessity for interpolating and increasing the resolution of spherical signals in these applications. For example, spatial or temporal interpolation is required for building a fine-grained weather forecasting system \citep{lam2022graphcast, bi2022pangu}. Increasing the resolution of CMB data may discover new insights about the early universe~\citep{bennett1996four, spergel2003first, ade2014planck}.

To this end, recent works \citep{esteves2022generalized, grattarola2022generalised, koestler2022intrinsic, schwarz2023modality, huang2023compressing} have investigated CNRs based on coordinate-based multilayer perceptrons (MLPs). To featurize the spherical coordinates, they use the sinusoids from the Cartesian coordinate system \citep{schwarz2023modality, huang2023compressing}, spectral embeddings of the discretized manifold \citep{grattarola2022generalised, koestler2022intrinsic}, or the spherical harmonics with fixed frequency components \citep{esteves2022generalized}. However, these methods suffer from the inherent limitation of coordinate-based MLPs: lack of expressive power to approximate highly nonlinear signals to their finest details. 

For Euclidean data, recent works \citep{chabra2020deep, muller2022instant, martel2021acorn, liu2020neuralsparsevoxel, peng2021neuralbody} have resolved this issue by using hybrid neural representations that combine feature-grid structures with the MLP. Their key idea is to decompose the input space into grid structures and assign learnable parameters as positional features for each point in the grid. Then the positional features can largely fluctuate accordingly with the unconstrained point-wise parameters. However, extending the existing hybrid neural representation to spherical data is non-trivial since they rely on grid structures and interpolation algorithms specific to Euclidean domains, e.g., sparse voxel grids and Euclidean distance-based interpolation, that are highly redundant or ill-defined for spherical data.

\textbf{Contribution.} In this work, we develop a novel hybrid neural representation for spherical data (HNeR-S) that combines spherical feature-grids with MLP. Our main idea is to use a hierarchy of coarse- to fine-grained spherical feature-grids to generate positional features of a spherical coordinate at multiple frequencies. We focus on scientific data: weather and climate data as well as CMB data. We provide an overview of HNeR-S in \cref{fig:overview}.

To compute the positional feature for an input point, we assign parameters to the grid point in the spherical feature-grid and apply a spherical interpolation algorithm to output the feature for the input position. Especially, we choose to align the feature-grid structure with the inherent data structure. This leads to designing our method with (1) an equirectangular feature-grid for the weather and climate data and (2) a hierarchical equal area iso-latitude pixelization~(HEALPix) feature-grid for the CMB data.






We thoroughly evaluate our algorithm on weather and climate \citep{hersbach2018era5} and CMB \citep{ade2014planck} data for super-resolution, regression, temporal interpolation, and compression tasks. Furthermore, we implement and compare with five CNR baselines \citep{mildenhall2021nerf, sitzmann2020implicit, esteves2022generalized, saragadam2023wire, tancik2020fourier} applicable to spherical data. 




To summarize, our contributions are as follows:
\begin{itemize}
    \item We propose a novel hybrid neural representation for spherical data (HNeR-S), which tailors the existing hybrid neural representations for spherical data using spherical feature-grids and interpolation algorithms.
    \item We empirically evaluate our algorithms on climate and weather data as well as CMB data. The results demonstrate our superiority over five INR baselines on super-resolution, regression, temporal interpolation, and compression tasks.
\end{itemize}

\section{Related Work}
\textbf{Coordinate-based neural representations (CNRs).} 
CNRs aim to learn a mapping from coordinates to the signal values via neural networks. The choice of positional encodings and activation functions are important for CNRs to capture high-frequency information of the underlying signal \citep{rahaman2019spectral}. To this end, sinusoidal positional encoding \citep{tancik2020fourier} and activation function \citep{sitzmann2020implicit} have been proposed for Euclidean data, e.g., images and videos. \citet{saragadam2023wire} proposed the Gabor wavelet activations to improve the performance and the robustness of sinusoidal encodings for the image domain. \citet{tancik2020fourier} analyzed the benefits of positional encoding to learn high-frequency functions via neural tangent kernel theory~\citep{jacot2018neural}.

\textbf{CNRs for non-Euclidean data.} Recently, researchers have also considered generalizing CNRs to non-Euclidean data \citep{esteves2022generalized, grattarola2022generalised, koestler2022intrinsic, schwarz2023modality, huang2022compressing}. A simple approach is to lift the two-dimensional sphere into three-dimensional Euclidean space so that the Cartesian coordinates of the sphere can be used as inputs to the existing three-dimensional CNRs \citep{schwarz2023modality, huang2023compressing}. To incorporate the geometry of the data, researchers considered using the eigenfunctions of the Laplace-Beltrami operator for general manifolds \citep{grattarola2022generalised, koestler2022intrinsic}. For spherical data, this corresponds to using spherical harmonics to featurize the spherical coordinates~\citep{esteves2022generalized}.

\textbf{Learning weather and climate data.} 
Recently, weather and climate data have gained popularity as an application of machine learning \citep{yuval2020stable}, which is attributed to its significant impact on climate prediction, mitigation, and adaptation. Employing neural networks for weather and climate data tasks has shown impressive results across a variety of tasks, e.g., super-resolution \citep{wang2021fast, yang2022statistical}, temporal modeling \citep{wang2018esrgan, stengel2020adversarial}, and compression \citep{dupont2022coin++, huang2023compressing}. Neural network architectures like spherical convolutional neural networks \citep{2018spherical} and spherical Fourier neural operators \citep{2023spherical} have been tailored for this domain.

\textbf{Learning cosmic microwave background~(CMB) data.} CMB data encapsulates the thermal radiation left over from the Big Bang, the origin of the universe. Machine learning has been applied to predict the posterior distribution of the cosmological parameters \citep{hortua2020parameter} as well as signal recovery and dust cleaning \citep{caldeira2019deepcmb, wang2022recovering, casas2022cenn}. A potential application of machine learning for CMB data is to increase resolution of the data, which has mainly been progressed from the physical development of new sensors \citep{bennett1996four, spergel2003first, ade2014planck}.

\section{Method}
\begin{figure}
    \centering
    \includegraphics[width=0.47\textwidth]{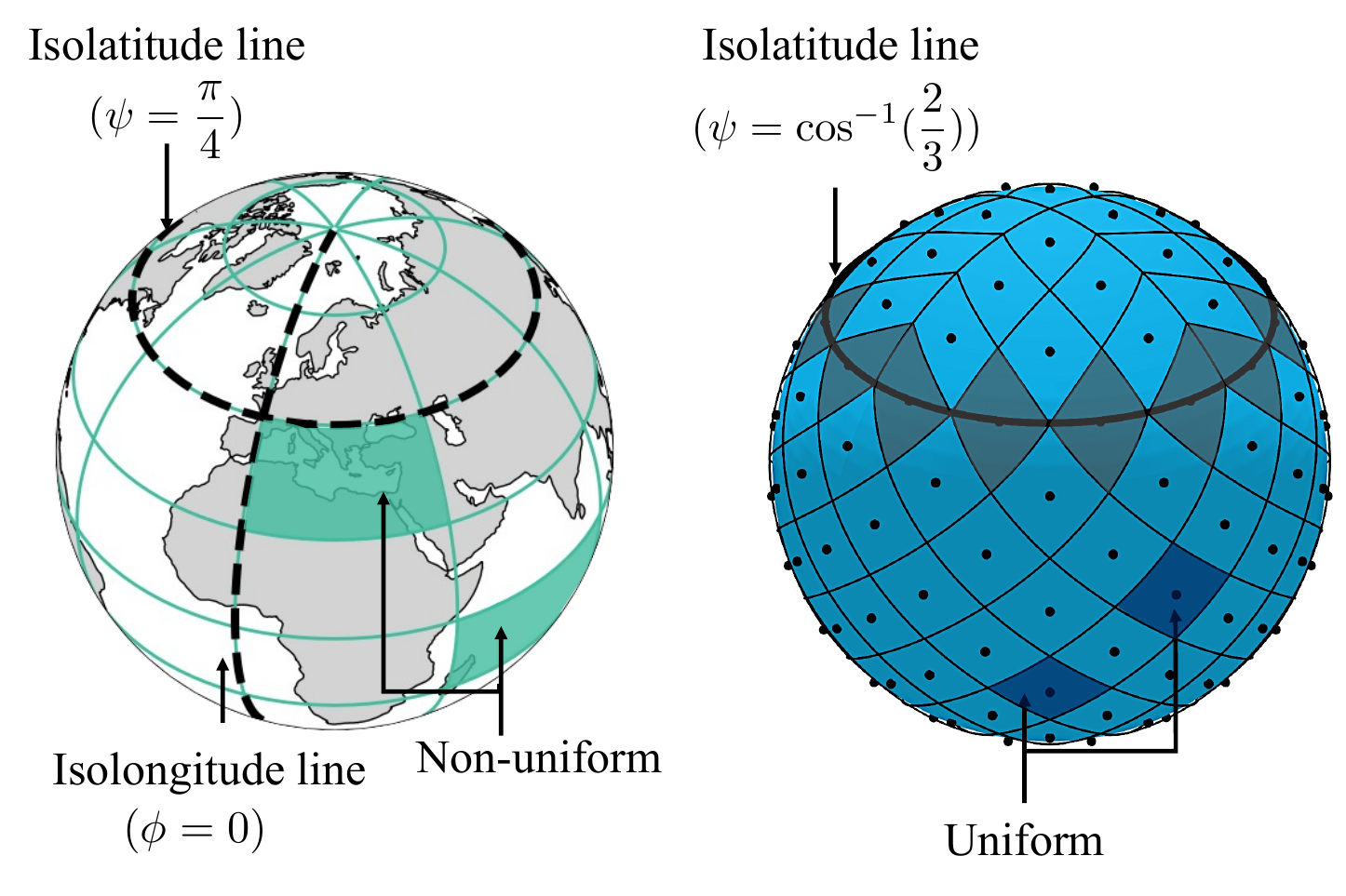}
    \caption{\textbf{Overview of equirectangular and HEALPix grid.} Note that the uniformity of feature-grid refers to the consistency in the area covered by each unit cell within the grid. }
    \label{fig:grid}
\end{figure}

In this section, we introduce our hybrid neural representations for spherical data (HNeR-S). We first elaborate on the general grid-agnostic framework (\cref{subsec:general}) and provide detailed descriptions for the equirectangular and HEALPix feature-grids (\cref{subsec:equi,subsec:heal}). We provide a brief overview of each grid in \cref{fig:grid}.

\subsection{General Framework}\label{subsec:general}
We formulate our problem as learning a mapping from spherical coordinates to the target signal. To this end, our HNeR-S employs a hierarchy of spherical feature grids and a bilinear spherical interpolation algorithm to output the positional feature of the input coordinate. Then HNeR-S applies a multi-layered perception (MLP) on the positional feature to predict the target signal. We provide an overview of HNeR-S in \cref{fig:overview}. 

\textbf{Problem definition.} To be specific, we consider learning some ground-truth spherical signal $f(\mathbf{x}_{\text{in}}) \in \mathbb{R}^{D_{\text{out}}}$, where the input coordinate $\mathbf{x}_{\text{in}}=(\psi, \phi)$ is a pair of latitude $\psi \in [-\frac{1}{2}\pi, \frac{1}{2}\pi]$ and longitude $\phi \in [0, 2\pi)$. To approximate this signal, we train a neural network $f_{\theta}$ on a dataset of tuples~$(\mathbf{x}_\text{in}, \mathbf{y})$ with ground truth label $\mathbf{y} = f(\mathbf{x}_\text{in})$.

\textbf{Overall architecture.} 
Our HNeR-S makes a prediction $\hat{\mathbf{y}}=f_{\theta}(\mathbf{x}_{\text{in}})$ from an input point $\mathbf{x}_{\text{in}}$ as follows:
\begin{equation*}
    f_{\theta}(\mathbf{x}) = \operatorname{MLP}(\mathbf{Z}(\mathbf{x}_{\text{in}})),
\end{equation*}
where $\operatorname{MLP}(\cdot)$ denotes the MLP and $\mathbf{Z}(\mathbf{x}_{\text{in}}) \in \mathbb{R}^{D\times L}$ 
denotes the concatenation of $D$-dimensional positional features constructed from $L$ levels of feature-grids:
\begin{equation*}
    \mathbf{Z}(\mathbf{x}_{\text{in}}) = \operatorname{Concat}(\mathbf{z}^{(1)}(\mathbf{x}_{\text{in}}),\ldots, \mathbf{z}^{(L)}(\mathbf{x}_{\text{in}})),
\end{equation*}
where $\operatorname{Concat}(\cdot)$ denotes concatenation of elements and the positional feature $\mathbf{z}^{(\ell)}(\mathbf{x}_{\text{in}}) \in \mathbb{R}^{D}$ is constructed from the $\ell$-th level feature-grid for $\ell=1,\ldots, L$. We assign a higher level $\ell$ for a finer grid. 

\textbf{Positional feature.} The $\ell$-th positional feature $\mathbf{z}^{(\ell)}(\mathbf{x}_{\text{in}})$ is constructed by interpolating the positional features assigned to grid points $\mathcal{V}^{(\ell)}$ in the $\ell$-th feature-grid. We let $\mathbf{x}_{i}$ denote the coordinate of a grid point $i \in \mathcal{V}^{(\ell)}$. Different grid types, e.g., equirectangular and HEALPix grids, correspond to a different organization of position $\mathbf{x}_{i}^{(\ell)}$ for each grid point~$i$. We also let $\mathbf{z}_{i}^{(\ell)} \in \mathbb{R}^{D}$ denote the associated positional feature, which is also a parameter to train for our framework. 

Then given an input point with position $\mathbf{x}_{\text{in}}$, the HNeR-S constructs the $\ell$-th positional feature $\mathbf{z}^{(\ell)}(\mathbf{x}_{\text{in}})$ by interpolation of neighboring points~$\mathcal{N}^{(\ell)}(\mathbf{x}_{\text{in}}) \subseteq \mathcal{V}^{(\ell)}$ in the feature-grid:
\begin{equation*}
    \mathbf{z}^{(\ell)}(\mathbf{x}_{\text{in}}) = \operatorname{Interpolate}(\mathbf{x}_{\text{in}}, \{(\mathbf{x}_{i}^{(\ell)}, \mathbf{z}_{i}^{(\ell)}):i \in \mathcal{N}^{(\ell)}(\mathbf{x}_{\text{in}})\}).
\end{equation*} 
The neighborhood structure of the equirectangular and HEALPix grid allows our HNeR-S to use bilinear spherical interpolation based on the latitude and the longitude. 

\textbf{Neighborhood structure.}
We consider neighborhood functions that output $\mathcal{N}^{(\ell)}(\mathbf{x}_{\text{in}})\subseteq \mathcal{V}^{(\ell)}$ as a set of four points, denoted by~$\mathbf{x}_{i,j} = (\psi_{i,j}, \phi_{i,j})$ associated with positional feature $\mathbf{z}_{i,j}^{(\ell)}$ for $i, j \in \{1,2\}$. Importantly, we consider grids that allow the construction of the neighborhood as pairs of isolatitude points and let $\psi_{1} = \psi_{1,1}=\psi_{1,2}, \psi_{2} = \psi_{2,1}=\psi_{2,2}$ without loss of generality. 

Furthermore, the neighbors bound the input point by a latitude interval $\Psi = [\psi_{1,1}, \psi_{1,2}]$ and two longitude intervals~$\Phi_{1}, \Phi_{2}$. When a neighbor of the input point does not overlap with the prime meridian ($\phi=0$), the longitude intervals are defined similarly, e.g., $\Phi_{1} = [\phi_{1,1}, \phi_{1,2}]$. Otherwise, the longitude intervals are set by $\Phi_{1} = [\phi_{1,1}, 2\pi] \cup [0, \phi_{1,2}]$ and $\Phi_{2}$ is set similarly.

\textbf{Bilinear spherical interpolation.} 
Given the neighbor as pairs of isolatitude points, our framework employs bilinear interpolation using the longitude and the latitude to compute the associated feature. To be specific, given an input point $\mathbf{x}_{\text{in}}=(\psi, \phi)$, we compute $\mathbf{z}^{(\ell)}(\psi, \phi)$ via two steps: (1) computing $\mathbf{z}^{(\ell)}(\psi_{1}, \phi)$ and $\mathbf{z}^{(\ell)}(\psi_{2}, \phi)$ via isolatitude interpolation and (2) computing $\mathbf{z}^{(\ell)}(\psi, \phi)$ via interpolation along the longitude between features computed in (1).

Precisely, the interpolation along the isolatitude neighbors $\mathbf{x}_{1, 1} = (\psi_{1}, \phi_{1,1}), \mathbf{x}_{1, 2}=(\psi_{1}, \phi_{1,2})$ is defined as follows:
\begin{align*}
    \mathbf{z}^{(\ell)}(\psi_{1}, \phi) 
    = \lambda \mathbf{z}_{1,1}^{(\ell)} + (1-\lambda) \mathbf{z}_{1,2}^{(\ell)},
\end{align*}
where $\mathbf{z}^{(\ell)}_{1,1} = \mathbf{z}^{(\ell)}(\psi_{1}, \phi_{1,1})$ and $\mathbf{z}^{(\ell)}_{1,2} = \mathbf{z}^{(\ell)}(\psi_{1}, \phi_{1,2})$ are learnable parameters. For the interpolation weight $\lambda$, the algorithm sets~$\lambda=\frac{d(\phi, \phi_{1,1})}{d(\phi_{1,1}, \phi_{1,2})}$ where $d(\cdot, \cdot)$ is the difference in the longitude for isolatitude points, e.g., $d(\phi_{1,1}, \phi_{1,2}) = \phi_{1,2} - \phi_{1,1}$ if the interval $\Phi_{1}$ does not overlap with the prime meridian. One can also compute $\mathbf{z}^{(\ell)}(\psi_{2}, \phi)$ similarly using $\mathbf{z}^{(\ell)}_{2,1}$ and $\mathbf{z}^{(\ell)}_{2,2}$. 

Finally, the algorithm computes the final positional feature by interpolation along the latitude:
\begin{equation*}
    \mathbf{z}^{(\ell)}(\psi, \phi) =  \mu \mathbf{z}^{(\ell)}(\psi_{1}, \phi)
    + (1-\mu) \mathbf{z}^{(\ell)}(\psi_{2}, \phi),
\end{equation*}
where the interpolation weight is set to $\mu = \frac{\psi - \psi_{1}}{\psi_{2} - \psi_{1}}$.

\subsection{Equirectangular Feature-Grid}\label{subsec:equi}

\begin{figure}[t]
    \centering
    \includegraphics[width=0.98\linewidth]{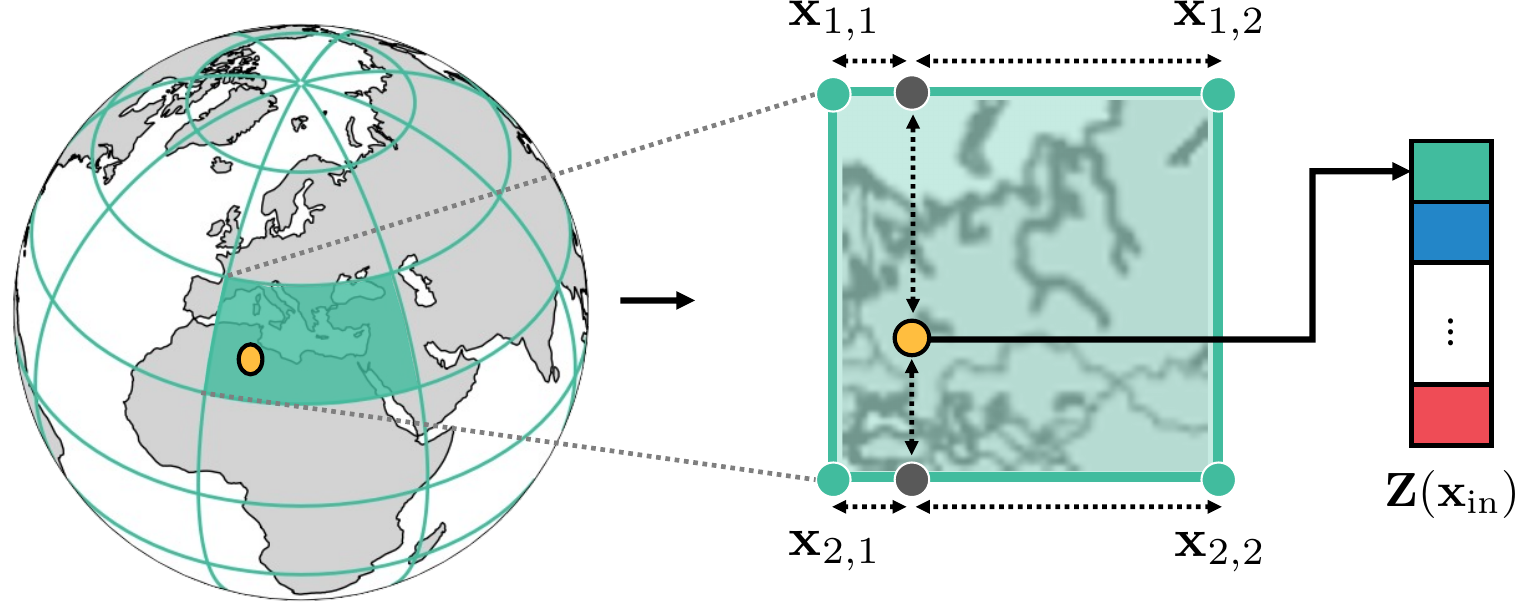}
    \caption{\textbf{Neighborhood structure of the equirectangular grid.} The yellow point indicates the input point $\mathbf{x}_{\text{in}}$ and green points indicate the neighborhood grid points $\mathcal{N}^{(\ell)}(\mathbf{x}_{\text{in}})$.}
    \label{fig:equi}
    \vspace{-0.1in}
\end{figure}
\begin{figure}[t]
    \centering
    \includegraphics[width=\linewidth]{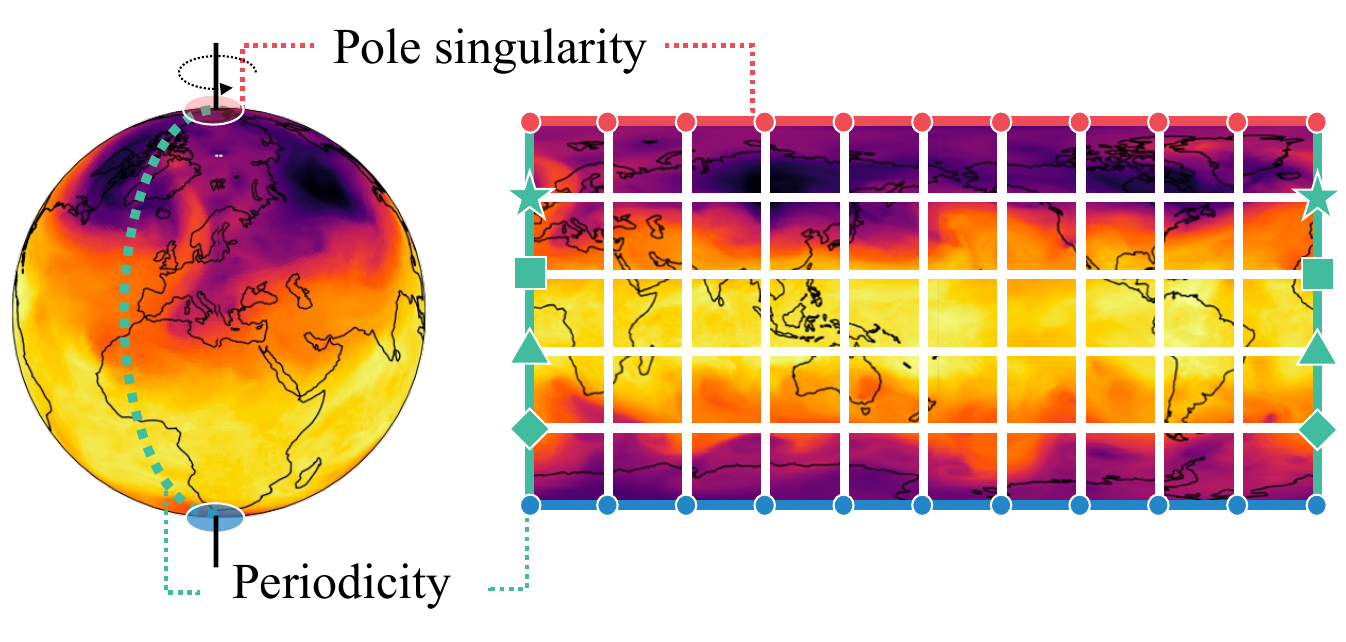}
    \caption{\textbf{Pole singularity and periodicity of equirectangular grid.} Different spherical coordinates can indicate the same point on a sphere, i.e., points at the North pole $(\psi=-\frac{\pi}{2})$, South pole $(\psi=\frac{\pi}{2})$, and the prime meridian ($\phi=0$ and $\phi=2\pi$). Our HNeR-S avoided assigning different parameters for such spherical coordinates. Points with the same marker and color share the associated parameters.}
    \label{fig:equi_const}
    \vspace{-0.2in}
\end{figure}
We use HNeR-S based on the equirectangular feature-grid for weather and climate data since the data is also distributed along an equirectangular grid. It is based on a projection that maps meridians (isolongitude lines) to vertical straight lines of constant spacing and circles of latitude (isolatitude lines) to horizontal straight lines of constant spacing. 
We index each point in the $\ell$-th grid by a tuple of integers $(n, m)$ with the position $\mathbf{x}_{n, m}^{(\ell)}$ described as follows:
\begin{align*}
    \mathbf{x}_{n,m}^{(\ell)} = (\psi_{n}^{(\ell)}, \phi_{m}^{(\ell)}) = \left(\pi\left(\frac{n}{N_{\text{lat}}^{(\ell)}} - \frac{1}{2}\right), 2\pi\frac{m}{N_{\text{lon}}^{(\ell)}}\right).
\end{align*}
The non-negative integers $n \leq N_{\text{lat}}^{(\ell)}$ and $m \leq N_{\text{lon}}^{(\ell)}$ index the point-wise latitude and longitude, respectively. The integers $N_{\text{lat}}^{(\ell)}, N_{\text{lon}}^{(\ell)}$ decide the latitude-wise and longitude-wise resolutions of the feature-grid. We choose them by:
\begin{align*}
N_\text{lat}^{(\ell)}=\lfloor \gamma^{\ell-1}\cdot N_\text{lat}^{(1)}\rfloor, \quad N_\text{lon}^{(\ell)}=\lfloor \gamma^{\ell-1}\cdot N_\text{lon}^{(1)}\rfloor,
\end{align*}
where $\gamma > 1$ is the scaling factor, $\lfloor\cdot\rfloor$ is the floor function, and $N_\text{lat}^{(1)}, N_\text{lon}^{(1)}$ are resolutions of the coarsest grid ($\ell=1$).

\textbf{Neighborhood structure.} 
Given an input point $\mathbf{x} = (\psi, \phi)$ as an input of the CNR, the neighborhood forms a unit cell of the feature-grid as follows:
\begin{align*}
    \mathcal{N}^{(\ell)}(\mathbf{x}) &= \{(n, m), (n, m+1), (n+1, m), (n+1, m+1)\}.
\end{align*}
Our framework determines the index $(n, m)$ to ensure the position $\mathbf{x}$ is inside the rectangle:
\begin{equation*}
    (n, m) = \left(\left\lfloor \left(\frac{\psi}{\pi}+\frac12\right)N^{(\ell)}_{\text{lat}} \right\rfloor, \left\lfloor \frac{\phi}{2\pi}N^{(\ell)}_{\text{lon}} \right\rfloor\right).
\end{equation*}
We describe the neighborhood structure in \cref{fig:equi}.

\textbf{Non-uniformity.} Since the equirectangular projection does not preserve area, the resulting equirectangular grid is irregular and the cell-wise areas are non-uniformly distributed. To incorporate this during training and evaluation, one needs to reweight each data point by the corresponding cell area~\citep{huang2023compressing}. The non-uniformity of the equirectangular grid is illustrated in \cref{fig:grid}.

\textbf{Pole singularity.} It is important to note that the equirectangular grid assigns multiple points on a single pole, i.e.,  $\mathbf{x}_{0,m} = \mathbf{x}_{0,m'}$ for any $m, m' < N_{\text{lon}}$. To incorporate this fact, we merge the associated parameters into a single parameter for both training and evaluation. We also remark on how the interpolation of our algorithm is well-defined even for the input points near the poles. We illustrate the pole singularity in \cref{fig:equi_const}.

\textbf{Periodicity. }The equirectangular projection of maps inherently introduces discontinuity, where the grid points sharing identical coordinates find themselves at the opposite ends of the map, on the left and right sides of the prime meridian, as described in \cref{fig:equi_const}. To address this issue, we merge the parameters corresponding to grid points at the same latitude from both ends of the map into a unified parameter.

\subsection{HEALPix Grid}\label{subsec:heal}

\begin{figure}[t]
    \centering
    \includegraphics[width=0.98\linewidth]{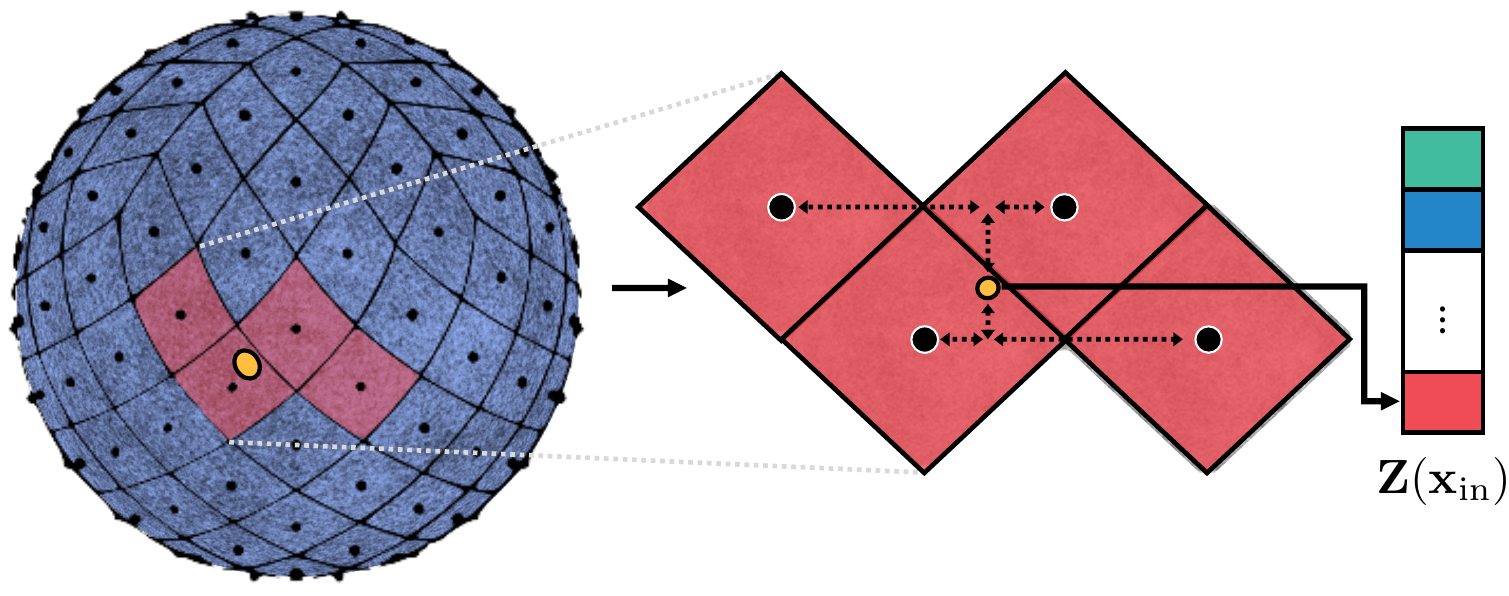}
    \vspace{-0.05in}
    \caption{\textbf{Neighborhood structure of the HEALPix grid.} The yellow point indicates the input point $\mathbf{x}$ and black points indicate the neighborhood grid points $\mathcal{N}^{(\ell)}(\mathbf{x}_\text{in})$.}
    \label{fig:healpix}
    \vspace{-0.1in}
\end{figure}

We consider feature-grids with hierarchical equal area isolatitude pixelation (HEALPix) structure for the CMB data \citep{gorski2005healpix}, since the data points are distributed according to the HEALPix grid structure. Notably, the data points are located in the center of the HEALPix grid cells. The grid structure uniformly partitions the sphere and the data points are uniformly distributed over the sphere. 

In particular, the HEALPix center points are located on $N_{\text{line}}$ lines of constant latitude, and each line (or azimuth) is uniformly divided by the points. The lines with latitude $\psi$ satisfying $|\cos \psi| \leq \frac23$ are divided into the same number of pixels $N_{\text{eq}}$. The remaining lines are located within the polar cap ($|\cos \psi| > \frac23$) and contain a varying number of pixels, increasing from line to line with increasing distance from the poles by one pixel within each quadrant. We refer to \citet{gorski2005healpix} for a detailed description.

The resolution of the grid is expressed by the parameter $N_{\text{side}}$, which denotes the number of divisions along the side of a base-resolution pixel that is needed to reach a desired high-resolution partition. One can observe that $N_{\text{line}}=4N_{\text{side}}-1$ and $N_{\text{eq}}=4N_{\text{side}}$. This results in $N_\text{pix}^{(\ell)} = 12 (N_\text{side}^{(\ell)})^2$ pixels and the corresponding center points. The resolution increases by the power of two for each level, i.e., we set $N_{\text{side}}^{(\ell)}=2^{\ell-1}$ for the $\ell$-th grid.

\textbf{Neighborhood structure.} We choose the neighborhood~$\mathcal{N}^{(\ell)}(\mathbf{x}_\text{in})$ in the HEALPix grid as the center of pixels that are adjacent to the pixel that includes the point $\mathbf{x}_\text{in}$. Furthermore, while a pixel in the HEALPix grid is adjacent to at most eight pixels, we choose the neighborhood as the center of the closest pixels on the two lines above and below the location. Such neighborhood allows the application of bilinear interpolation as described in \cref{subsec:general}. We illustrate the HEALPix neighborhood structure in \cref{fig:healpix}.

\section{Experiments}
In this section, we assess the performance of our two hierarchical hybrid neural representations designed for spherical data: the equirectangular grid and the HEALPix grid. To this end, we first validate our framework for climate data regarding four tasks: regression, super-resolution, temporal interpolation, and compression (\cref{subsec:exp1}). We also explore two tasks on CMB data with the HEALPix grid regarding super-resolution and regression (\cref{subsec:exp2}). We will release our code upon acceptance.

\begin{table}[t]
    \centering
    \caption{\textbf{Details on the datasets used for weather and climate (ERA5) and CMB (PR1) data.}}
    \label{tab:dataset_main}
    \vspace{0.1in}
    \resizebox{0.48\textwidth}{!}{
     \begin{tabular}{lcc}
        \toprule
         & {ERA5} & {PR1} \\
        \midrule
        {Grid} &  Equirectangular & HEALPix\\
        {Target} &  Geopotential, Temperature & CMB Temperature\\
        {Spatial Res.} & $0.25^\circ$, $0.50^\circ$, $1.00^\circ$ & $5 $ arcmin\\ 
        {Temporal Res.} & Daily, Weekly & -\\ 
        \bottomrule
      \end{tabular}
    }
    \vspace{-0.1in}
\end{table}

\textbf{Datasets.} Mainly, we utilized climate and CMB data for our experiment. Weather and climate data is gathered from ECMWF reanalysis 5th generation \citep[ERA5]{hersbach2018era5} archive where data can be directly downloaded by using Climate Data Storage API \citep{buontempo2020fostering}. To be specific, we gathered geopotential and temperature data in the year 2000 for super-resolution, regression, and temporal interpolation. For the compression task, we gathered geopotential data in the year 2016 following \citet{huang2023compressing}. The detailed description of datasets is in \cref{tab:dataset_main}.

Next, we use the CMB temperature data from Planck Public Data Release 1 (PR1) Mission Science Maps data at  NASA/IPAC Infrared Science Archive (IRSA)\footnote{\url{https://irsa.ipac.caltech.edu/data/Planck/release_1/all-sky-maps/}}. We use the version where missing values are determined by spectral matching independent component analysis \citep[SMICA]{delabrouille2003multidetector}. 

\textbf{Baselines.} On one hand, for super-resolution, regression, and temporal interpolation tasks, we compare our methods to five different encoding schemes that are applicable for learning spherical signals: logarithmically spaced sinusoidal positional encoding \citep[ReLU+P.E.]{mildenhall2021nerf}, sinusoidal representation networks \citep[SIREN]{sitzmann2020implicit}, wavelet implicit neural representation \citep[WIRE]{saragadam2023wire}, spherical harmonics implicit neural representation \citep[SHINR]{esteves2022generalized}, and Fourier feature networks \citep[FFN]{tancik2020fourier}.

On the other hand, for the compression task, we compare our methods to the state-of-the-art compression scheme without neural network \citep{liang2022sz3} and the recently proposed coordinate-based neural network for compression~\citep{huang2023compressing}.

\textbf{Experimental setting. }We conduct all experiments using a single RTX 3090 GPU. For the regression, super-resolution, and temporal interpolation experiments, we employ a consistent architecture, consisting of a 4-layer multilayer perceptron (MLP) with 256 units in each hidden layer. Detailed settings for the compression experiment are provided in \cref{subsec:exp1}.

\subsection{Weather and Climate Data}\label{subsec:exp1}

\begin{table}[t]
    \centering
    \caption{\textbf{Results of regression on weather and climate data.} The evaluation metric is weighted PSNR and the best metric is highlighted in \textbf{bold}.}
    \label{tab:weather_spatial_reg}
    \vspace{0.1in}
     \begin{tabular}{lcc}
        \toprule
        & Geopotential & Temperature \\
        \midrule
        ReLU+P.E. &  60.20 & 55.63\\
        SIREN &  55.10 & 55.20\\
        SHINR & 54.98 & 45.75\\ 
        WIRE & 65.93 & 59.81 \\ 
        FFN & 61.15 & 57.14 \\ 
        \midrule
        \textbf{HNeR-S (ours)} & \textbf{66.17} &  \textbf{63.45} \\ 
        \bottomrule
      \end{tabular}
\end{table}
\begin{figure}[t]
    \centering
    \captionof{table}{\textbf{Results of super-resolution on weather and climate data.} The evaluation metric is weighted PSNR and the best metric is highlighted in \textbf{bold}.}
    \label{tab:weather_spatial_super}
    \vspace{0.1in}
     \begin{tabular}{lcccc}
        \toprule
        & \multicolumn{2}{c}{Geopotential} & \multicolumn{2}{c}{Temperature} \\
        \cmidrule(lr){2-3} \cmidrule(lr){4-5}
         & $\times 2$ & $\times 4$ & $\times 2$ & $\times 4$ \\
        \midrule
        ReLU+P.E. & 61.71 & 58.07 & 54.17 & 51.35 \\
        SIREN & 60.52 & 57.64 &55.43 & 50.64 \\
        SHINR & 54.25 & 53.12  &43.00 & 42.27 \\ 
        WIRE & 59.67 & 55.34  &53.37 & 50.45  \\ 
        FFN & 62.04 & 56.62  &56.15 & 51.97 \\ 
        \midrule
        \textbf{HNeR-S (ours)} & \textbf{70.52} & \textbf{61.17}  & \textbf{61.28} & \textbf{53.93}  \\ 
        \bottomrule
      \end{tabular}
    
    
    \vspace{-0.1in}
    
    
\end{figure}

\begin{figure*}[t]
    \centering
    \scalebox{0.95}{
    \begin{tabular}[width=0.85\textwidth]{p{0.045\textwidth}cccccp{0.08\textwidth}}
    \toprule
         & Reference & RELU+P.E. & WIRE & FFN & \textbf{Ours} & \\
         \midrule
        \raisebox{-0.73\height}[0pt][0pt]{\includegraphics[height=8.6\normalbaselineskip,width=0.05\textwidth]{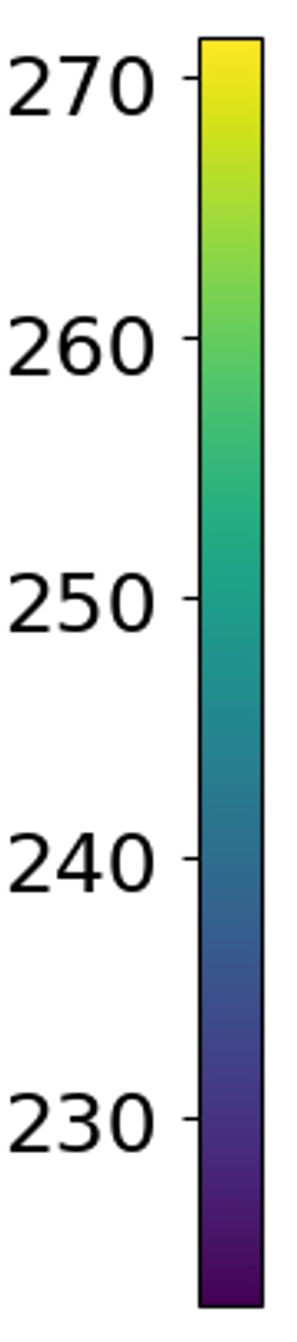}}
         & \begin{minipage}{0.115\textwidth}
             \includegraphics[width=\linewidth]{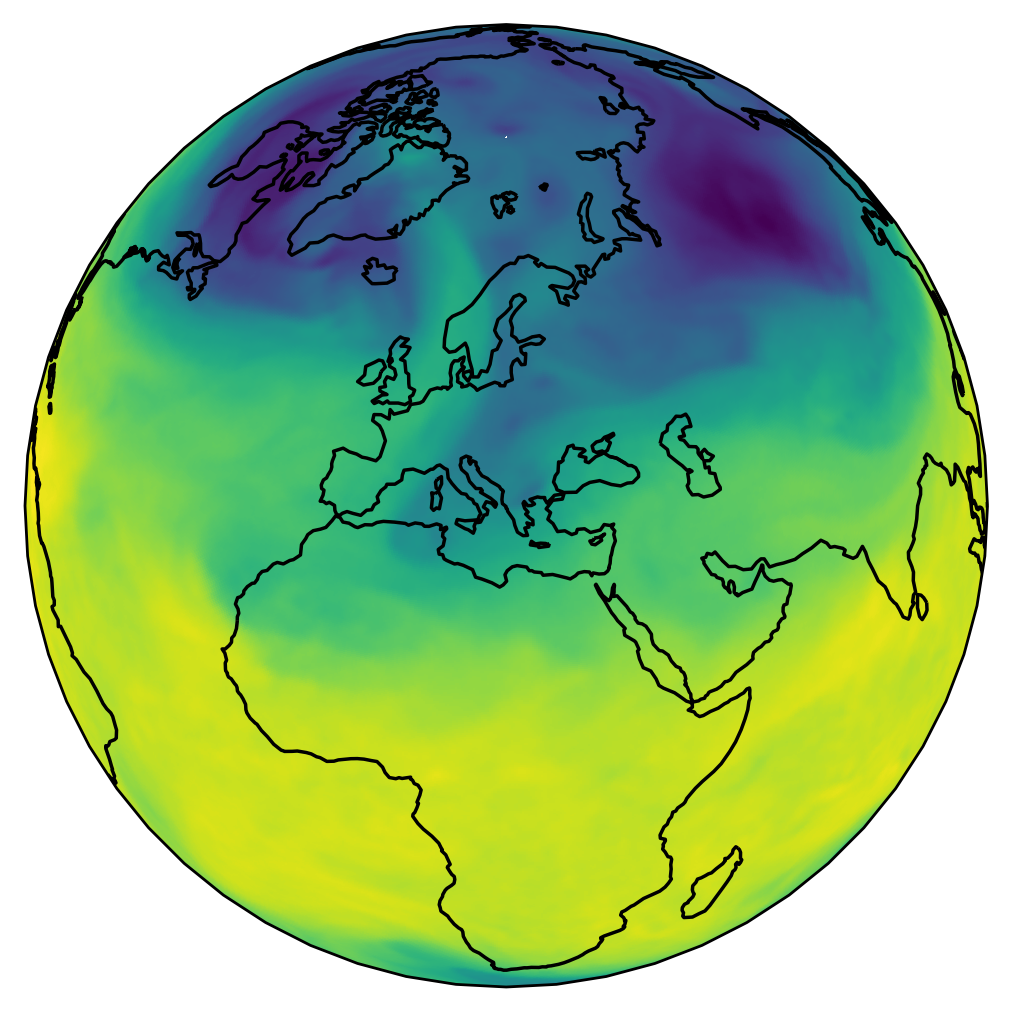}
         \end{minipage} &
              \begin{minipage}{0.115\textwidth}
             \includegraphics[width=\linewidth]{figure/weather_x2/sphere_relu_pred}
         \end{minipage} &
              \begin{minipage}{0.115\textwidth}
             \includegraphics[width=\linewidth]{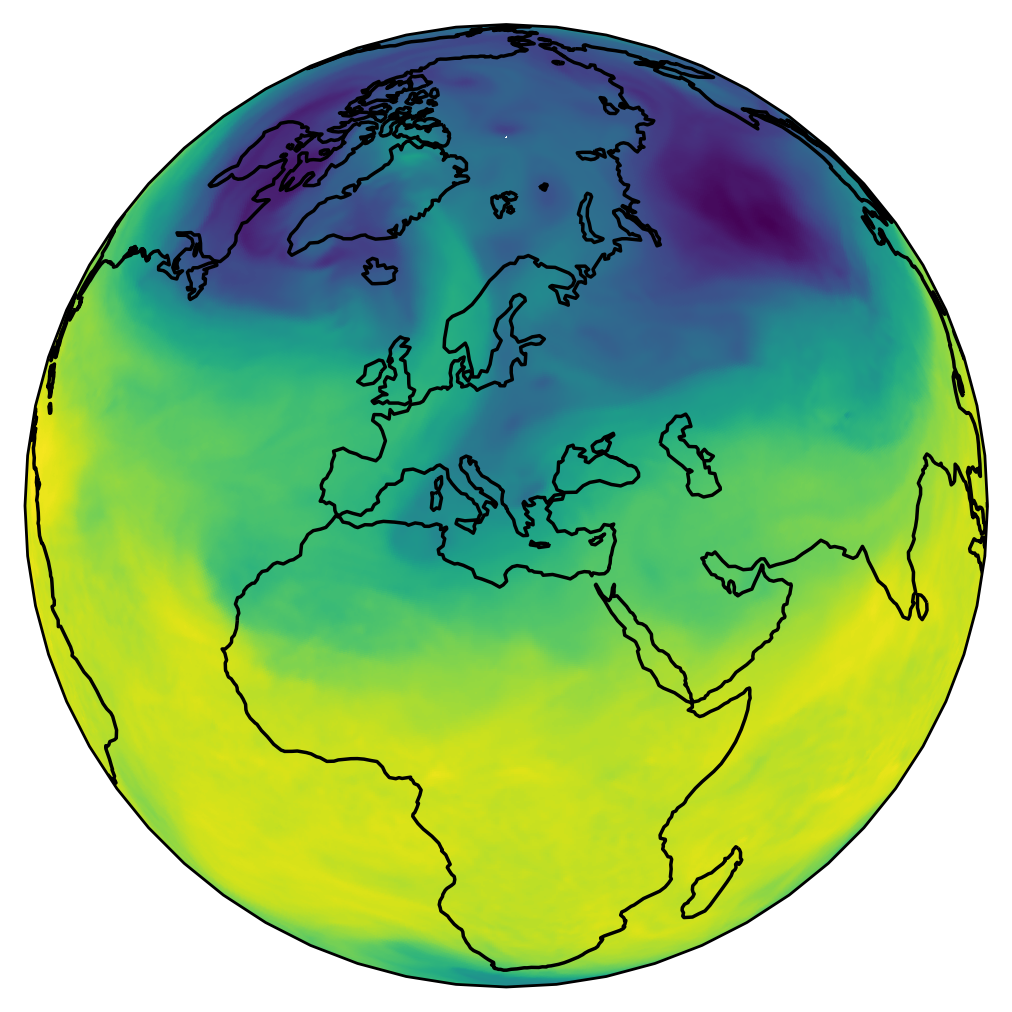}
         \end{minipage} &
              \begin{minipage}{0.115\textwidth}
             \includegraphics[width=\linewidth]{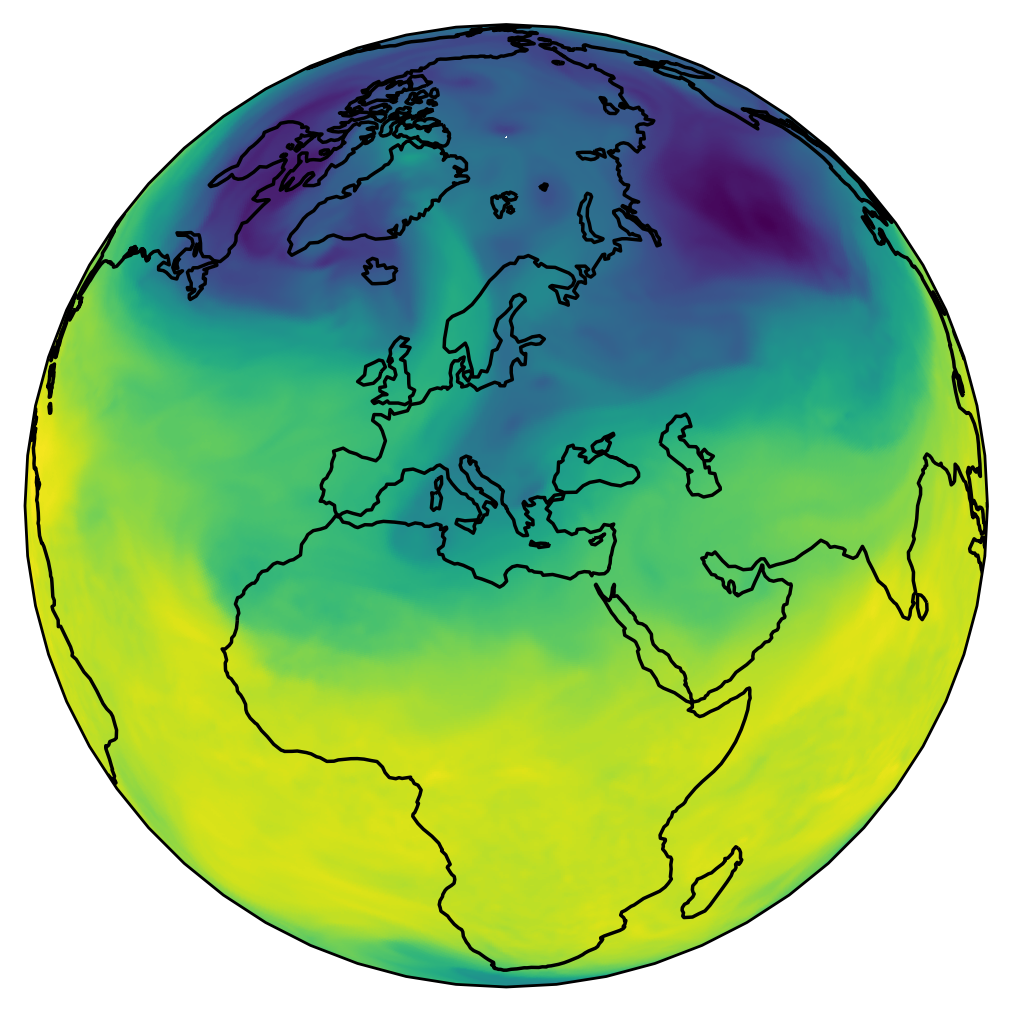}
         \end{minipage} &
              \begin{minipage}{0.115\textwidth}
             \includegraphics[width=\linewidth]{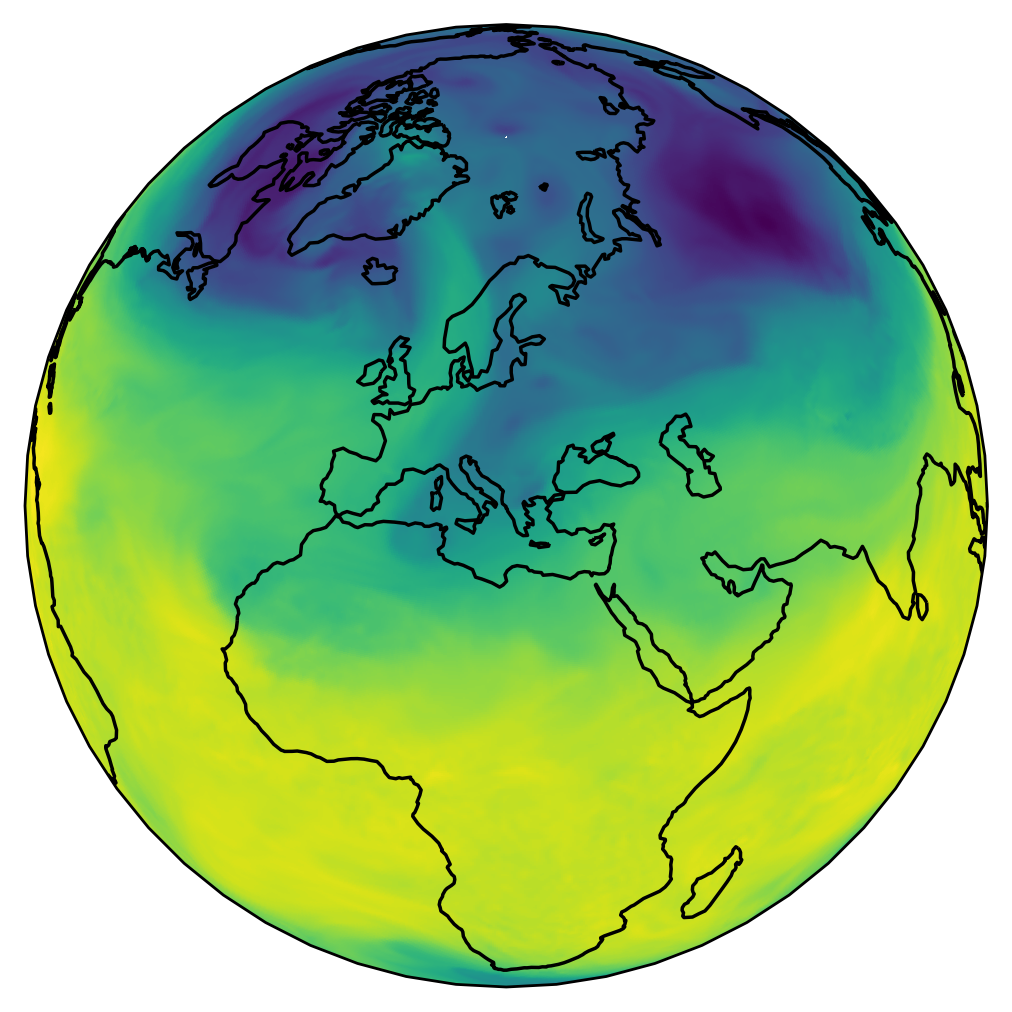}
         \end{minipage} &
         \raisebox{-0.725\height}[0pt][0pt]{\includegraphics[height=8.6\normalbaselineskip,width=0.05\textwidth]{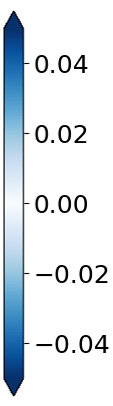}}
         \\
         
         & &
              \begin{minipage}{0.115\textwidth}
             \includegraphics[width=\linewidth]{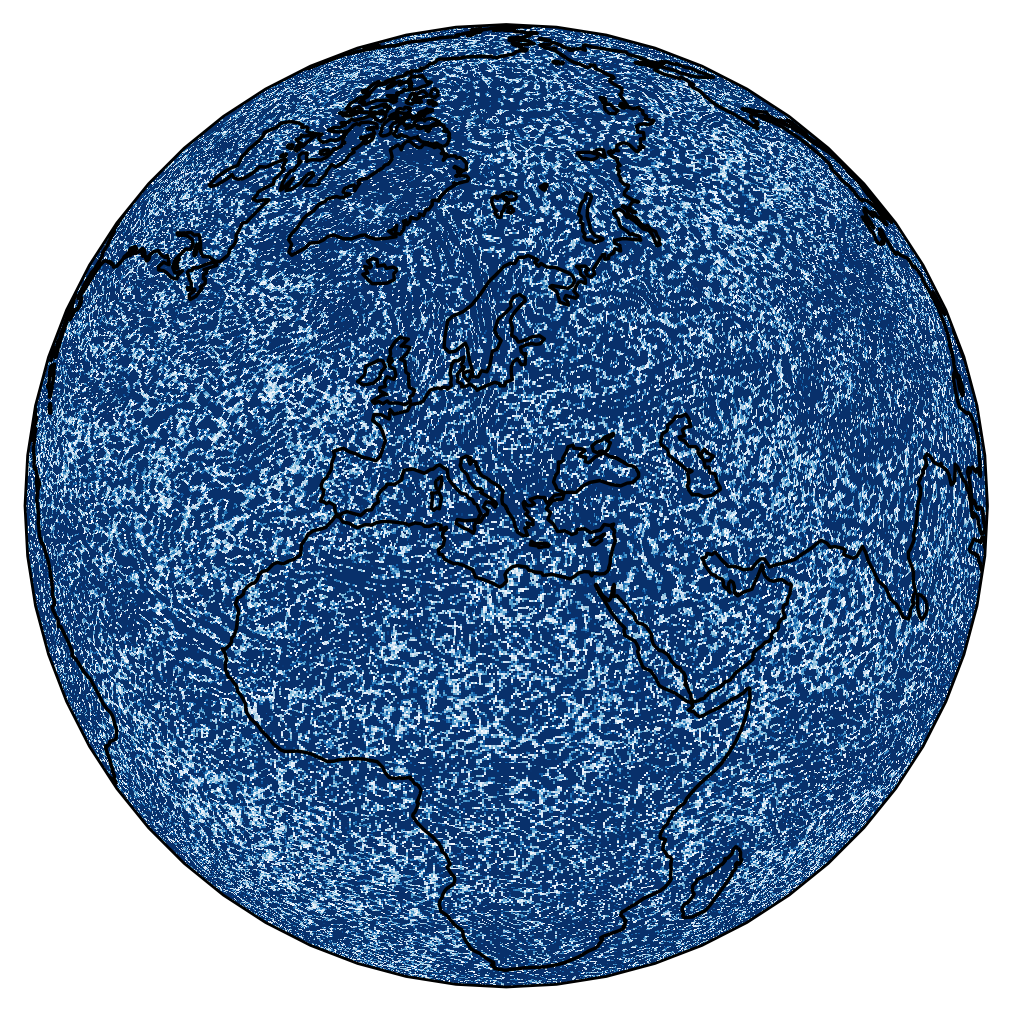}
         \end{minipage} &
              \begin{minipage}{0.115\textwidth}
             \includegraphics[width=\linewidth]{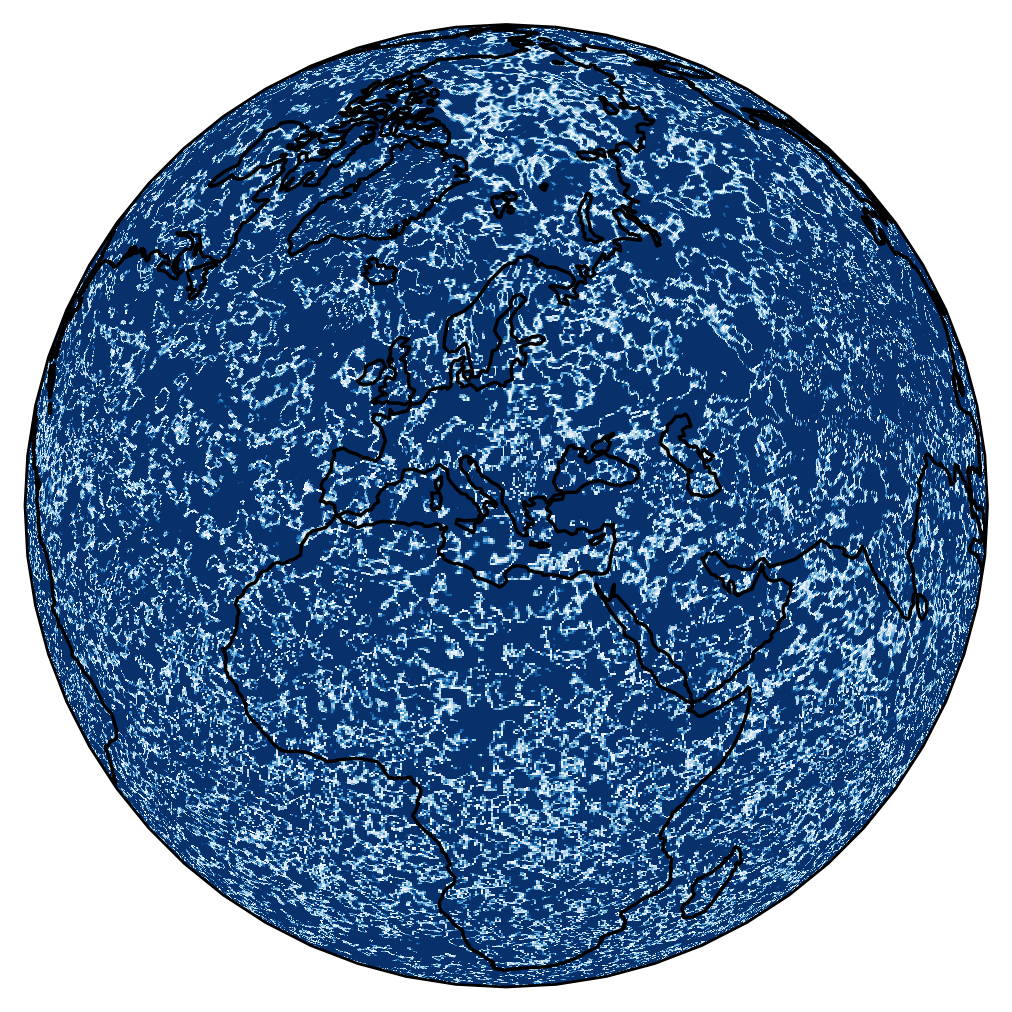}
         \end{minipage} &
              \begin{minipage}{0.115\textwidth}
             \includegraphics[width=\linewidth]{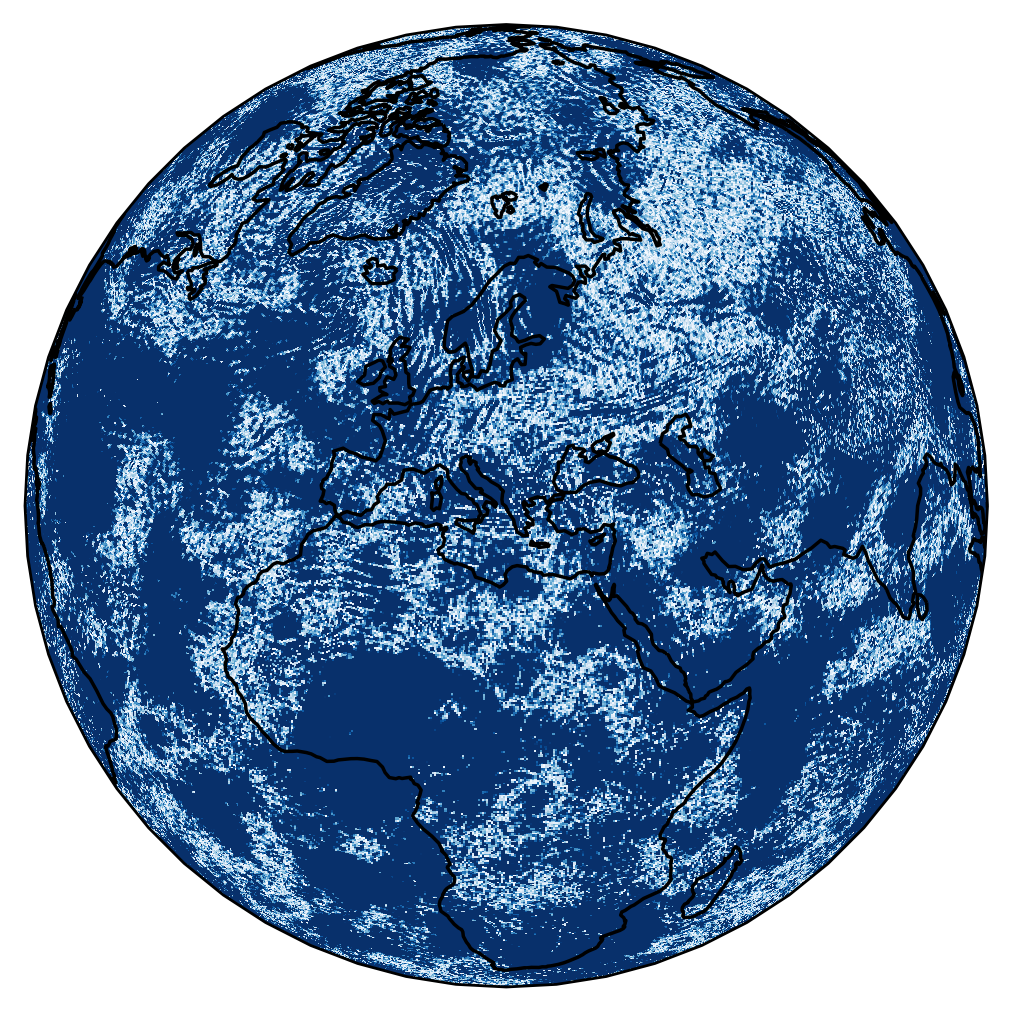}
         \end{minipage} &
              \begin{minipage}{0.115\textwidth}
             \includegraphics[width=\linewidth]{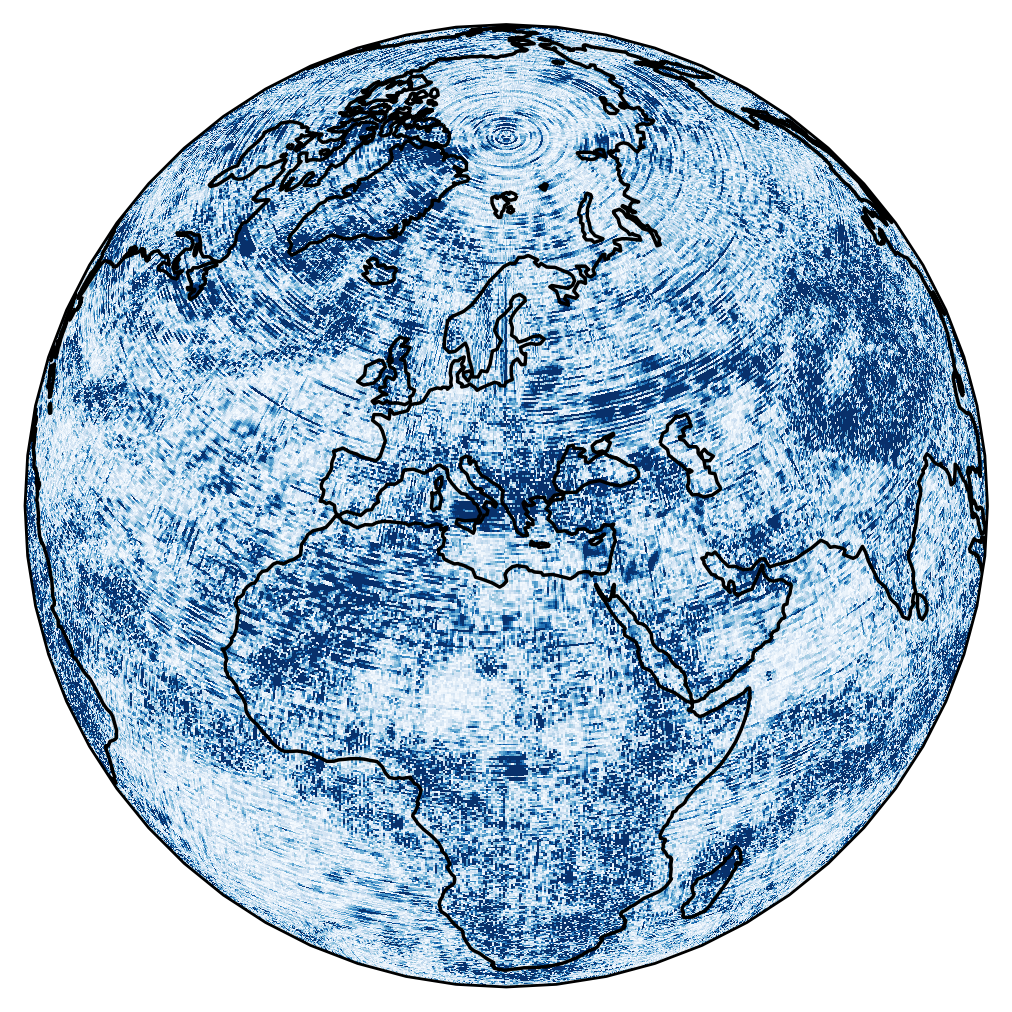}
         \end{minipage} \\
         \bottomrule
    \end{tabular}
    }
    \caption{\textbf{Prediction (top) and error maps (bottom) for $\times2$ super-resolution in climate and weather temperature data.}}
    \label{fig:vis_weather}
\end{figure*}

In this section, we evaluate our model on weather and climate data for $\times2$ and $\times4$ super-resolution, regression, temporal interpolation, and compression. The climate data is often represented using an equirectangular grid (i.e., a regular latitude-longitude grid), where data distribution is concentrated around the poles. To address this unique challenge, we employ our hierarchical equirectangular feature-grid. In addition, every task on an equirectangular grid is evaluated by latitude-weighted metric, i.e., weighted RMSE, MAE, and PSNR \citep{huang2023compressing}. This assigns smaller weights to the poles, compensating for the irregularity of data sampled from the equirectangular grid.


\begin{figure*}
    \centering
    \includegraphics[width=\linewidth]{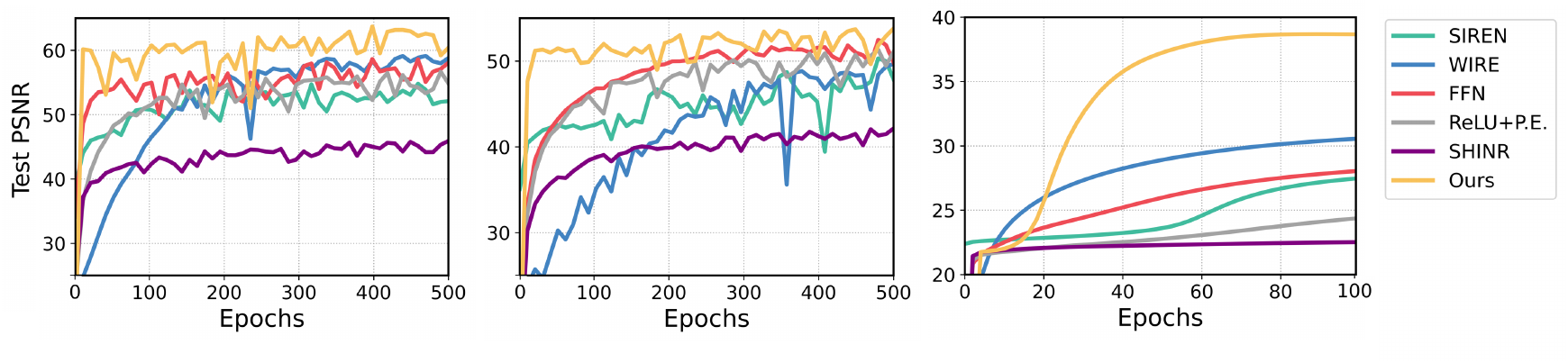}
    \caption{\textbf{Test PSNR curves.} These plots show the test PSNR curve for ($\times 4$) super-resolution (left) and regression (middle) for the temperature of weather and climate data and regression for CMB data (right). Our model converges to the best PSNR in the fastest manner.}
    \label{fig:psnr}
\end{figure*}

\textbf{Regression.} We first evaluate our framework for the regression task where only a portion of the grid is used for training and the rest is used for evaluation. We report the results in \cref{tab:weather_spatial_reg}. One can observe how our method consistently demonstrates superiority in PSNR performance and faster convergence over the baselines. We further visualize our results in \cref{fig:psnr}. The figure illustrates our method's accuracy in regression tasks, showcasing minimal errors across diverse geographical regions. 

In \cref{fig:psnr}, one can observe how our algorithm demonstrates significant improvement over the baselines. The improvement is more pronounced for predicting the temperature, where our HNeR-S achieves a 63.45 PSNR score whereas the second-best baseline (WIRE) achieves 59.81. 

\textbf{Super-resolution.} Next, we evaluate our method for the super-resolution task. To this end, we train the models on low-resolution grids and evaluate them on high-resolution grids. Precisely, we train the models on the resolution of 0.50$^\circ$ and $1.00^{\circ}$ evaluate them on the resolution of 0.25$^\circ$ for the $\times 2$ and $\times 4$ setting, respectively.

We report the results in \cref{tab:weather_spatial_super}. Here, one can observe how our algorithm demonstrates significant improvements over the baselines. We also visualize the results in \cref{fig:vis_weather}, which demonstrate our method's ability to reconstruct high-resolution details accurately, improving clarity and definition in climate data visualizations. 

\begin{table}[t]
    \centering
    \caption{\textbf{Performance for temporal interpolation on weather and climate data.} The best number is highlighted in \textbf{bold}.}
    \label{tab:weather_temporal}
    \vspace{0.1in}
     \begin{tabular}{lcccc}
        \toprule
        & \multicolumn{2}{c}{Geopotential} & \multicolumn{2}{c}{Temperature} \\
        \cmidrule(lr){2-3} \cmidrule(lr){4-5}
        & Daily & Weekly & Daily & Weekly  \\
        \midrule
        ReLU+P.E. & 28.43 & 23.92 & 27.12 & 24.35\\
        SIREN &\textbf{28.80} & 23.16 &27.26 & 23.34\\
        SHINR &  28.46 & 22.75 &  27.27 &22.90 \\
        WIRE &  28.66 & 22.99 & 27.07&23.18 \\
        FFN &  28.53 & 23.54 & 26.65 &23.73 \\ 
        \midrule
        \textbf{HNeR-S (ours)} & 28.79 & \textbf{23.97} & \textbf{27.55} & \textbf{24.50}\\
        \bottomrule
      \end{tabular}


\end{table}

\textbf{Temporal interpolation.} Finally, we evaluate the ability of our framework to interpolate between snapshots of the weather at different timesteps. This task employs two temporal resolutions: daily and weekly, with a total of 30 data snapshots sampled. Since this task requires the CNRs to condition on the timestep of the snapshot, we modify our algorithm and the baselines by incorporating the timestep $t$ as a concatenation of the positional feature, which serves as input for the MLP. 

The outcomes, detailed in \cref{tab:weather_temporal} and \cref{fig:psnr}, highlight our method's exceptional PSNR performance. The visualizations in \cref{fig:vis_weather} reveal our model's capability to accurately interpolate temporal changes.
\begin{table}[t]
    \centering
    \caption{\textbf{Results of compression on weather and climate data.} The best results of spatial compression models are highlighted in \textbf{bold}. $^{\dagger}$FFN-T and SZ3 results are from \citet{huang2023compressing}.}
    \label{tab:comp}
    \vspace{.1in}
    \resizebox{\linewidth}{!}{
\begin{tabular}{lcccccc}
    \toprule
    & \multicolumn{3}{c}{ $0.031 \leq $ BPP $\leq 0.080$} & \multicolumn{3}{c}{$0.099 \leq$ BPP $\leq 0.119$} \\
    \cmidrule(lr){2-4} \cmidrule(lr){5-7}
         & BPP  & WRMSE  & WMAE  & BPP  & WRMSE  & WMAE \\
         & $(\downarrow)$ & $(\downarrow)$ & $(\downarrow)$ & $(\downarrow)$ & $(\downarrow)$ & $(\downarrow)$ \\
         \midrule
    \multicolumn{7}{c}{Spatiotemporal Compression} \\
    \midrule
    SZ3$^{\dagger}$ & 0.080 & \phantom{0}698.5 & \phantom{0}574.9 & 0.115 & 493.9 & 408.8 \\
    FFN-T$^{\dagger}$ & 0.031 & \phantom{0}{143.1} & \phantom{0}{101.2} & 0.111 & {\phantom{0}73.6} & {\phantom{0}53.0} \\
    \midrule
    \multicolumn{7}{c}{Spatial Compression} \\
    \midrule
    SIREN & 0.063 & \phantom{0}669.5 & \phantom{0}462.1 & 0.119 & 616.0 & 404.8 \\
    WIRE & 0.069 & 2617.2 & 1433.3 & 0.099 & 688.7 & 478.6 \\
\midrule
\textbf{HNeR-S (ours)} & 0.057 & \textbf{\phantom{0}144.8} & \textbf{\phantom{00}99.4} & 0.114 & \textbf{124.0} & \textbf{\phantom{0}50.3} \\
\bottomrule
\end{tabular}}
\end{table}

\begin{table*}[t]
    \centering
    \scalebox{0.95}{
    \begin{tabular}[width=0.93\textwidth]{p{0.05\textwidth}cccccp{0.05\textwidth}}
    \toprule
         & Reference & RELU+P.E. & WIRE & FFN & \textbf{Ours} & \\
         \midrule
        \raisebox{-0.73\height}[0pt][0pt]{\includegraphics[height=8.6\normalbaselineskip,width=0.05\textwidth]{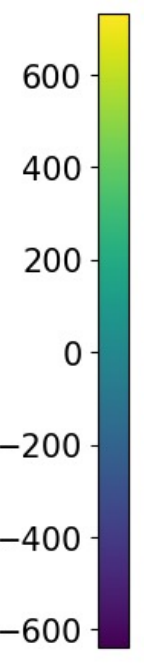}}
         & \begin{minipage}{0.15\textwidth}
             \includegraphics[width=\linewidth, height=1.9cm]{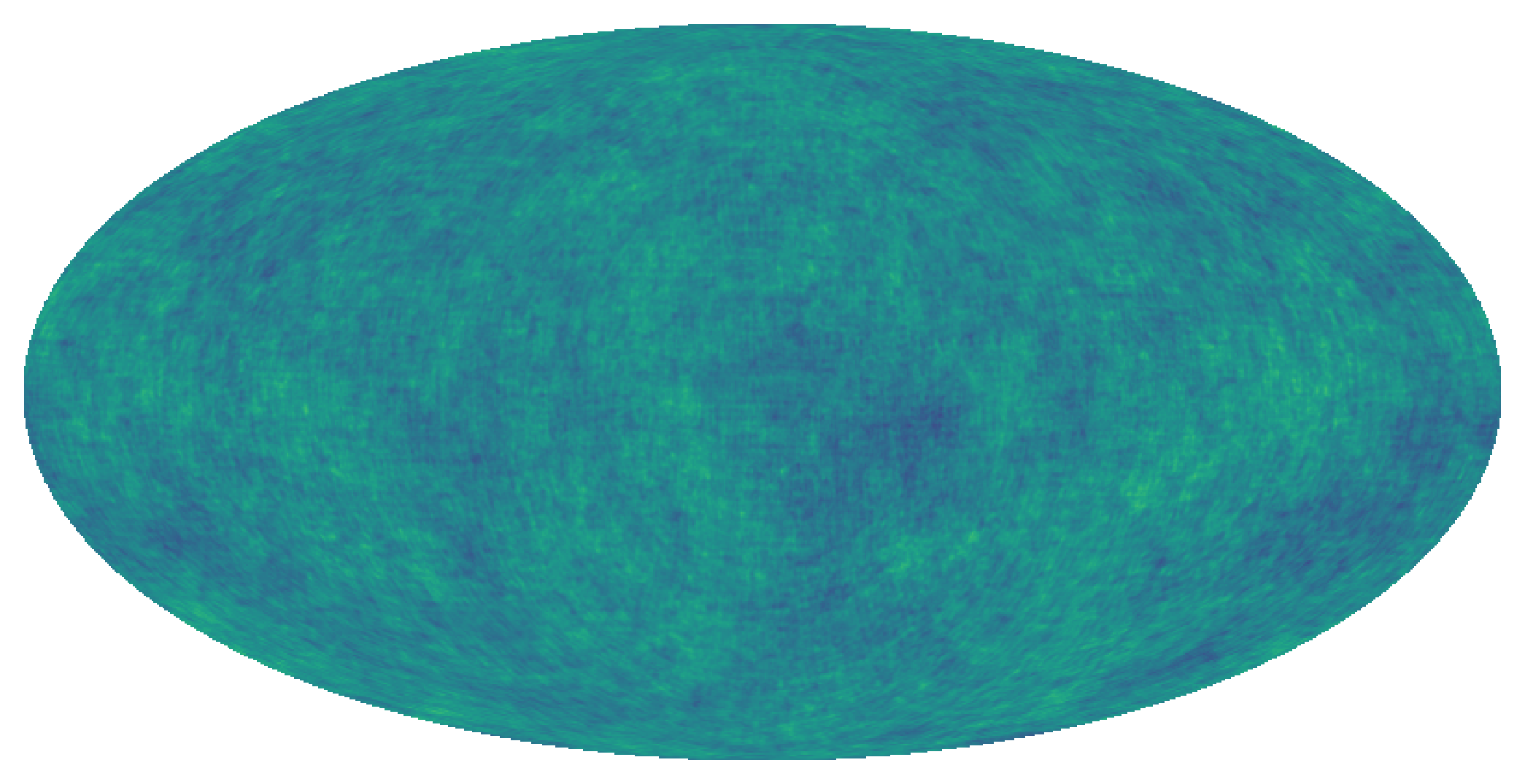}
         \end{minipage} &
              \begin{minipage}{0.15\textwidth}
             \includegraphics[width=\linewidth, height=1.9cm]{figure/cmb_reg/relu_pred.png}
         \end{minipage} &
              \begin{minipage}{0.15\textwidth}
             \includegraphics[width=\linewidth, height=1.9cm]{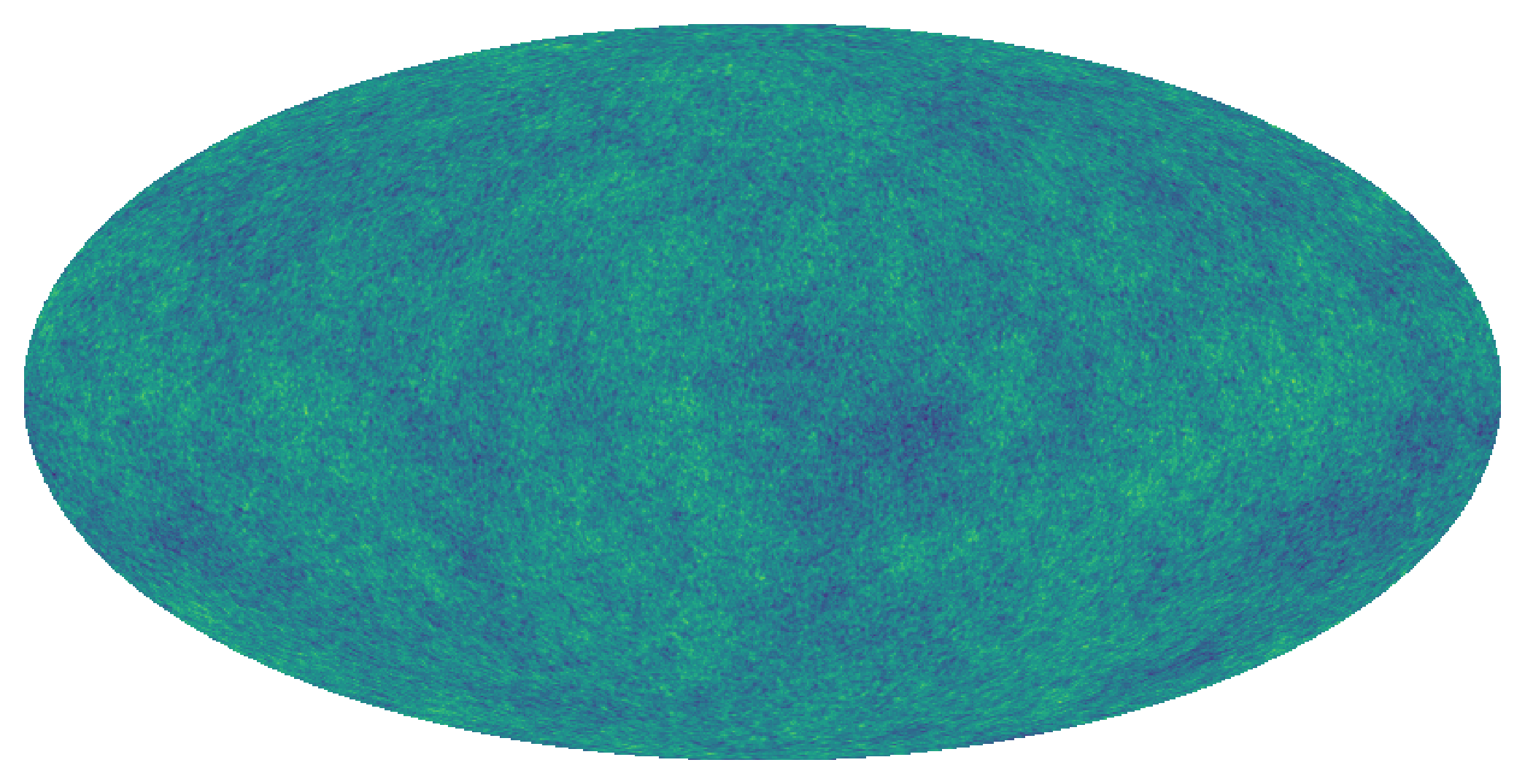}
         \end{minipage} &
              \begin{minipage}{0.15\textwidth}
             \includegraphics[width=\linewidth, height=1.9cm]{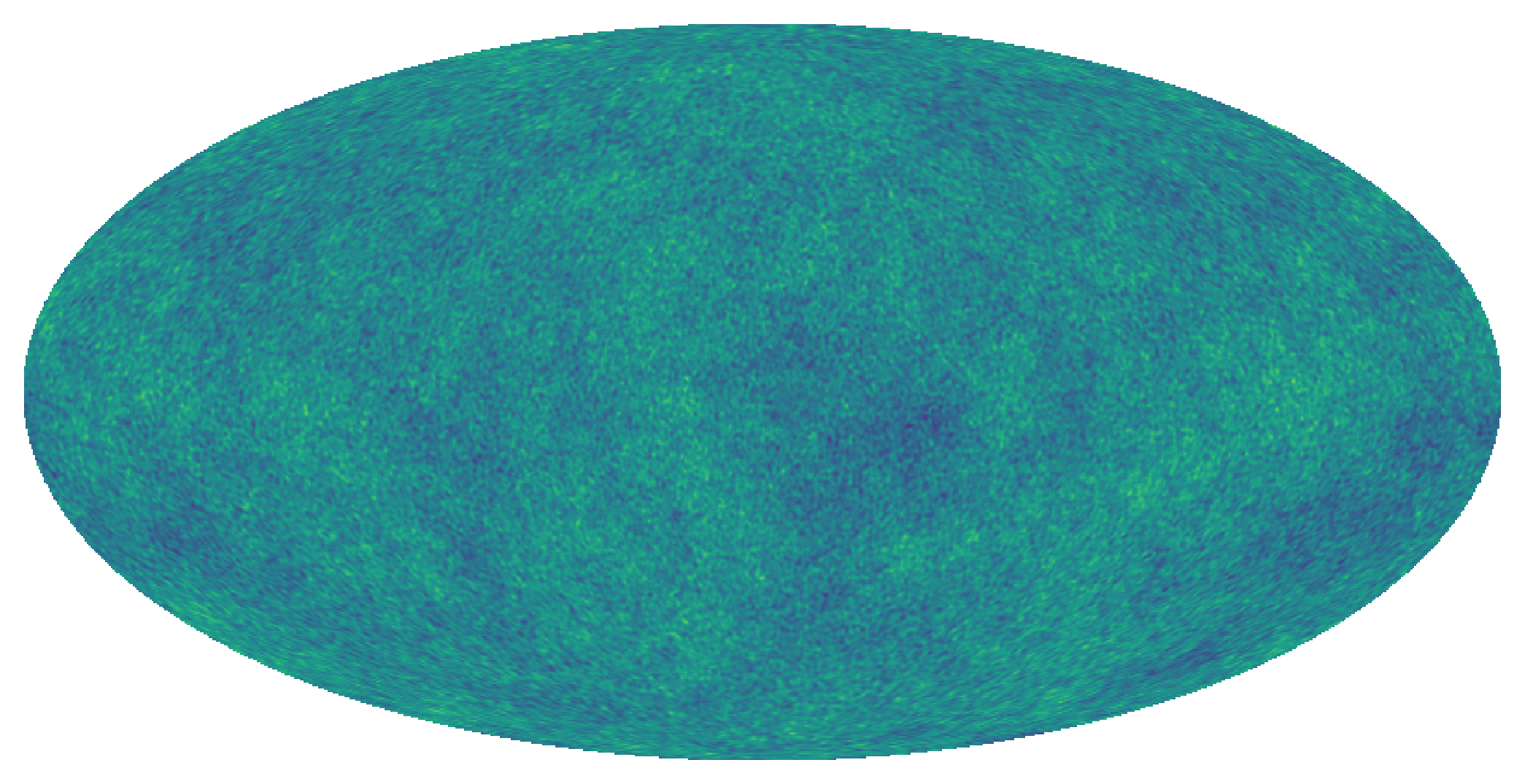}
         \end{minipage} &
              \begin{minipage}{0.15\textwidth}
             \includegraphics[width=\linewidth, height=1.9cm]{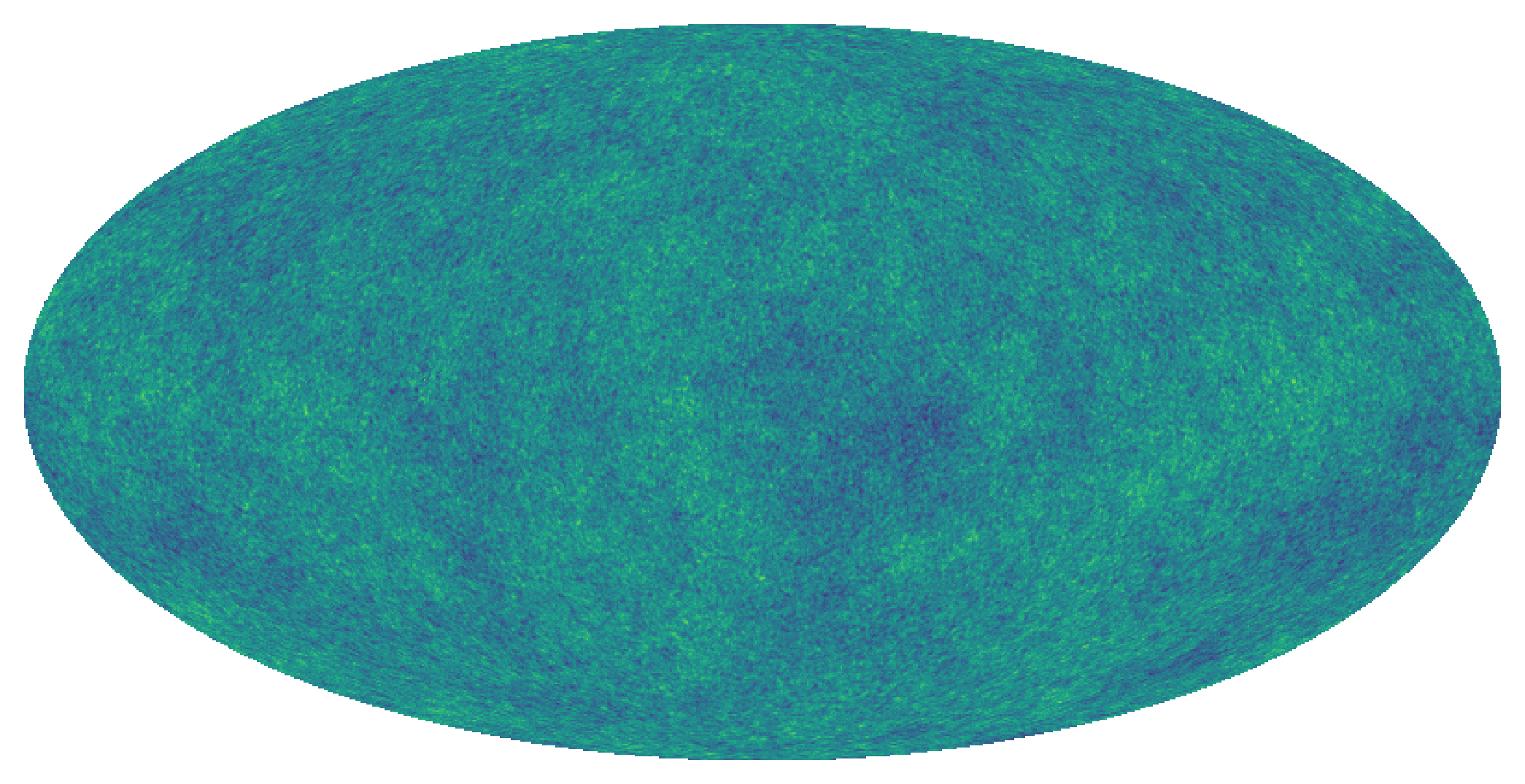}
         \end{minipage} &
         \raisebox{-0.725\height}[0pt][0pt]{\includegraphics[height=8.6\normalbaselineskip,width=0.05\textwidth]{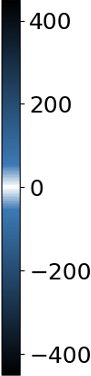}}
         \\
         
         & &
              \begin{minipage}{0.15\textwidth}
             \includegraphics[width=\linewidth, height=1.9cm]{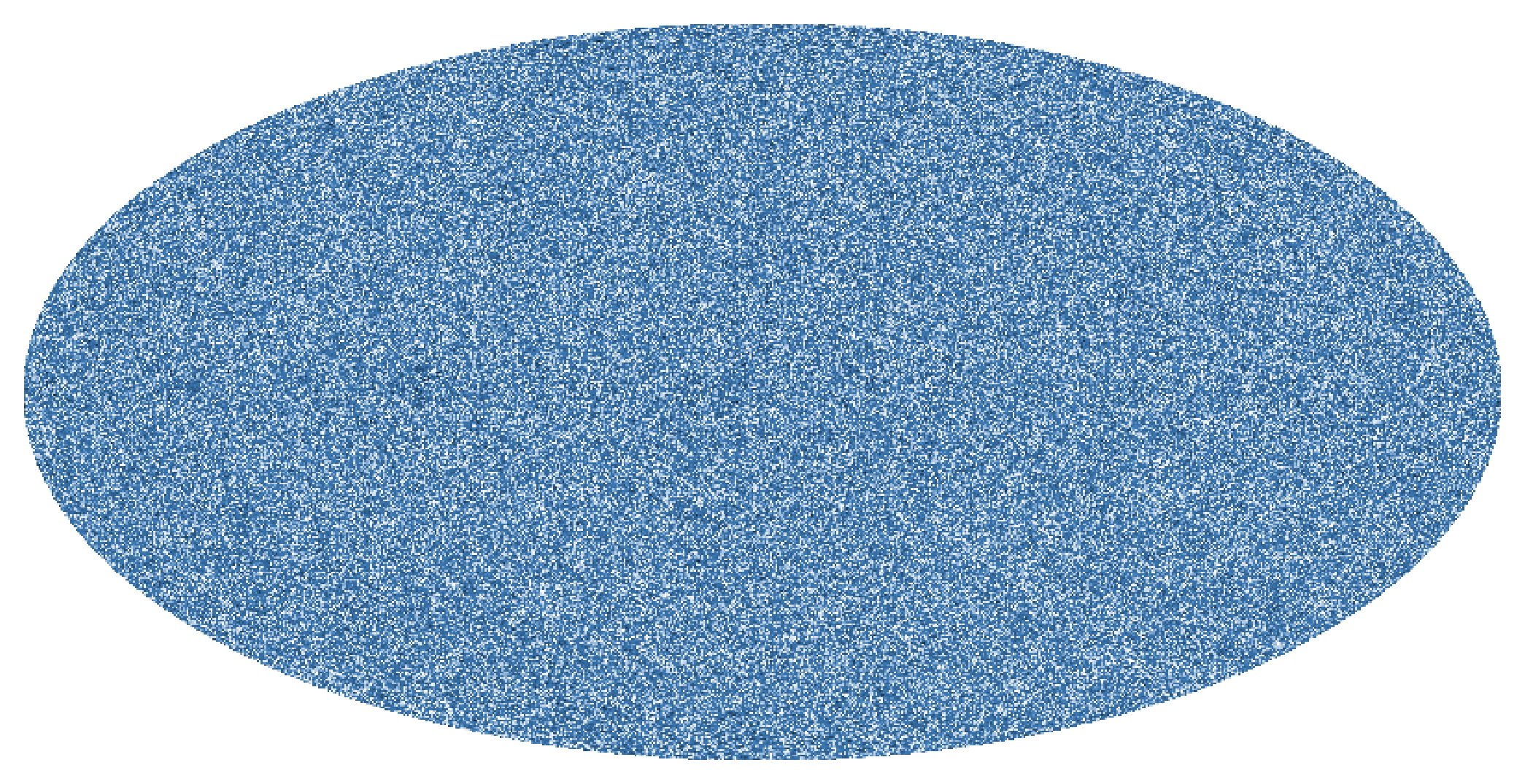}
         \end{minipage} &
              \begin{minipage}{0.15\textwidth}
             \includegraphics[width=\linewidth, height=1.9cm]{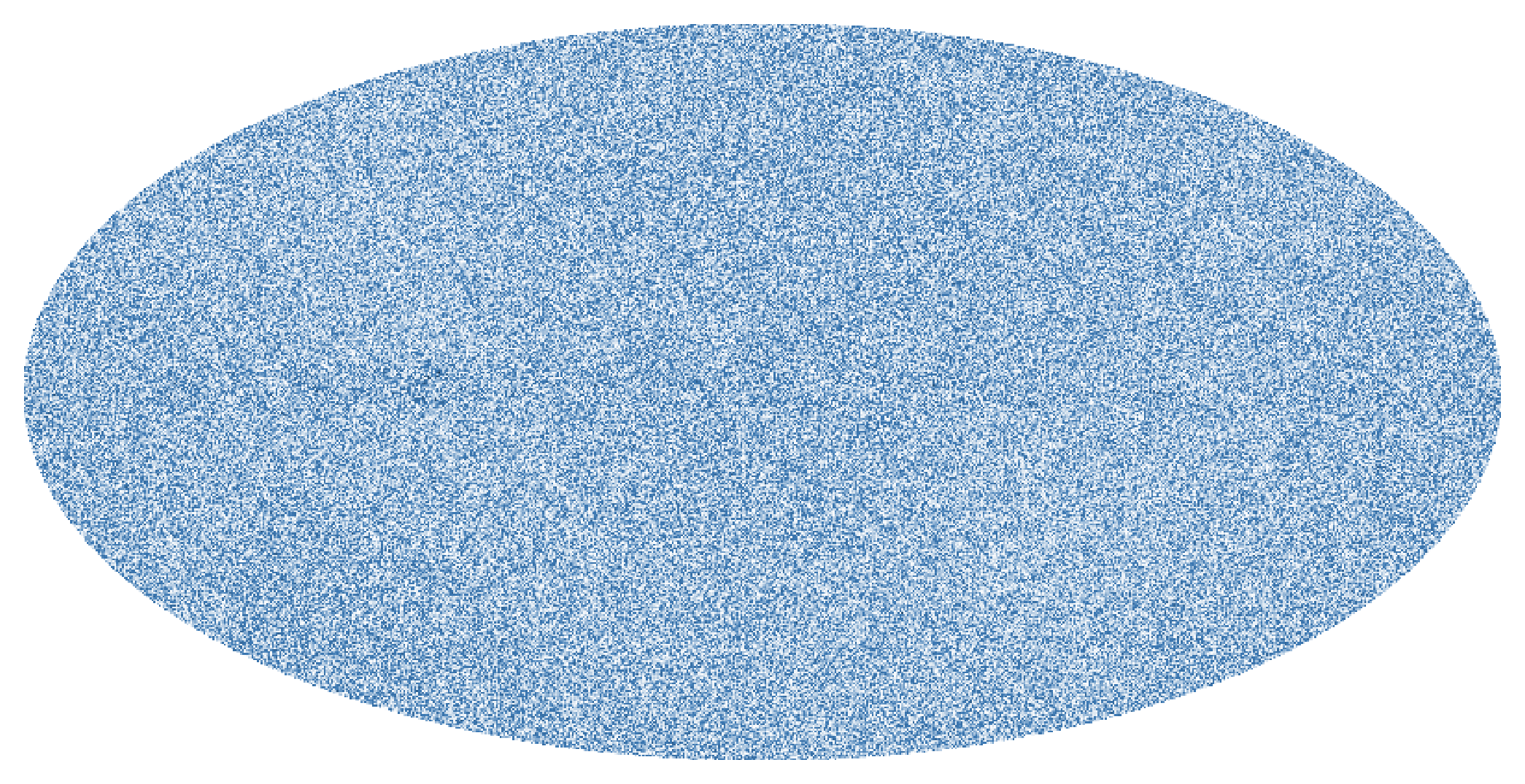}
         \end{minipage} &
              \begin{minipage}{0.15\textwidth}
             \includegraphics[width=\linewidth, height=1.9cm]{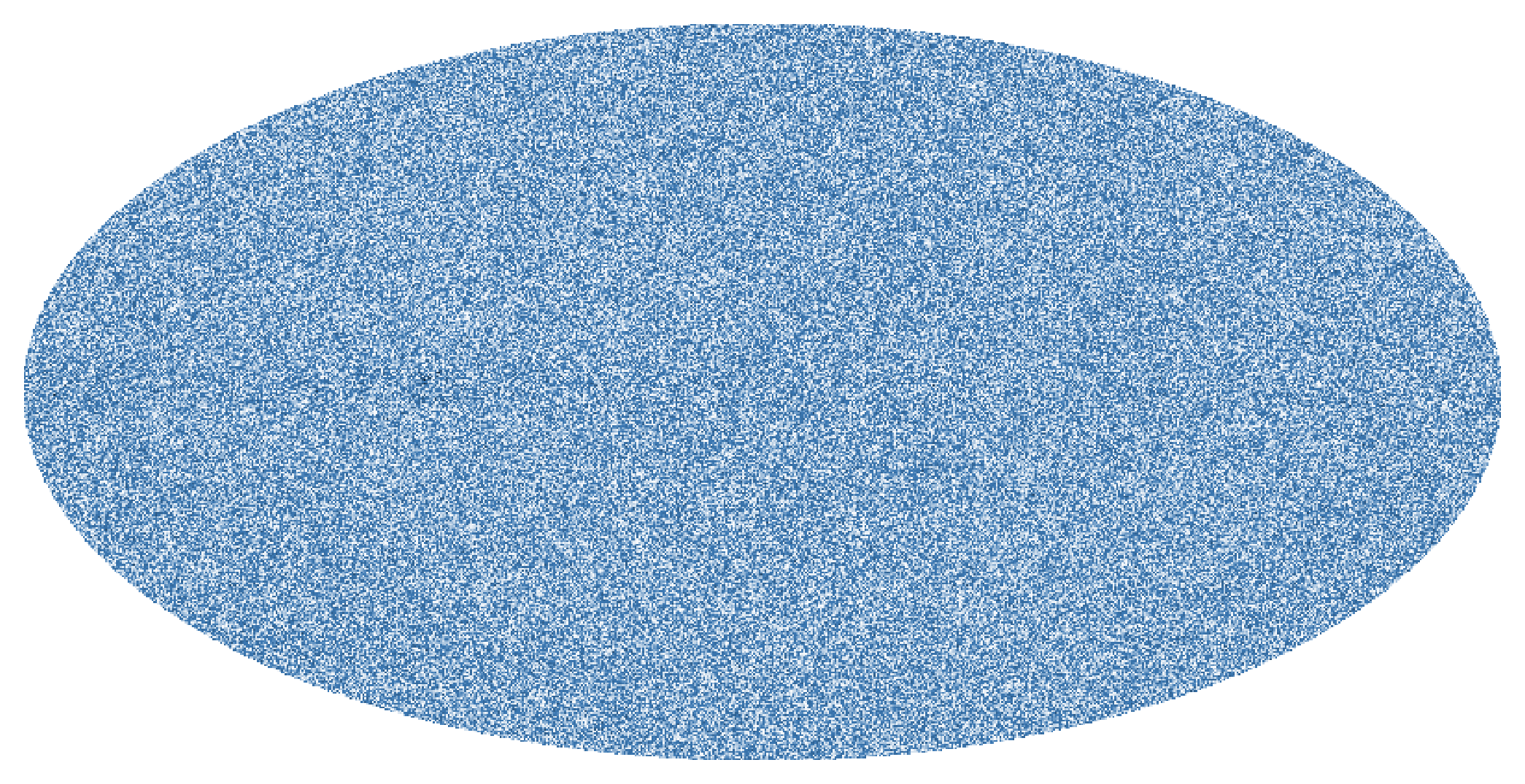}
         \end{minipage} &
              \begin{minipage}{0.15\textwidth}
             \includegraphics[width=\linewidth, height=1.9cm]{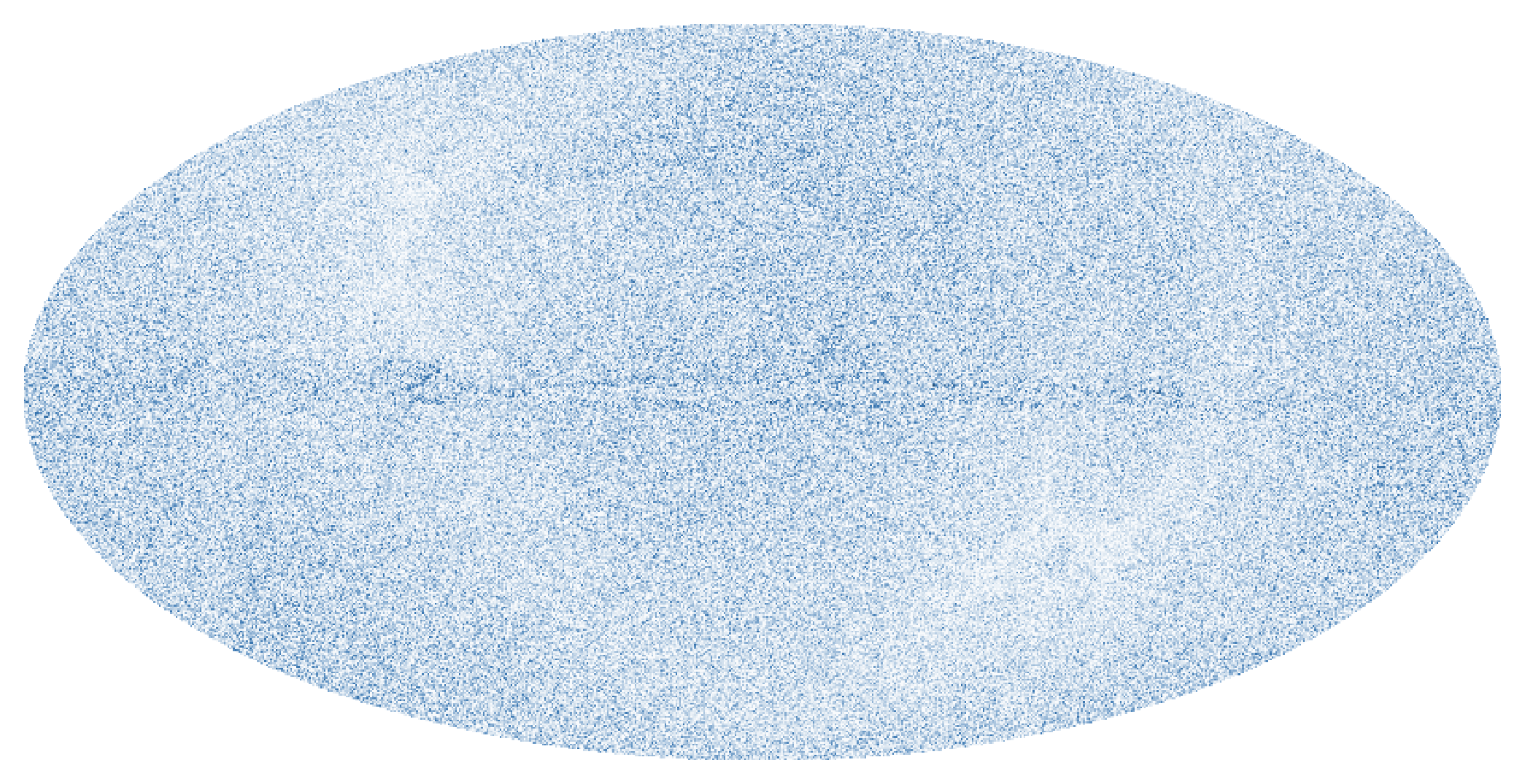}
         \end{minipage} \\
         \bottomrule
    \end{tabular}
    }
    \captionof{figure}{\textbf{Prediction (top) and error maps (bottom) for regression in CMB data.}}
    \label{fig:vis_cos}
\end{table*}

\textbf{Compression.} As an additional experiment we apply HNeR-S for compression, driven by the demand to store daily-growing high-resolution climate and weather datasets \citep{huang2023compressing}. To this end, we combine our method with the existing HNeR-based compression scheme \citep{ladune2023cool} to apply image-wise compression, i.e., we train one neural network for the compression of each image. As a non-machine learning baseline, we consider a modular composable SZ3 \cite{liang2022sz3} framework. We also compare with the Fourier feature network with temporal encoding \citep[FFN-T]{huang2023compressing} proposed for compression of climate and weather data. We note that FFN-T is a spatiotemporal compressor, i.e., it takes both spatial and temporal coordinate that trains a series of snapshots, while ours is a spatial compressor, i.e., it trains a network for each image. Additionally, we consider two  implicit neural network baselines including WIRE \citep{saragadam2023wire} and SIREN \citep{esteves2022generalized} for fair comparison on spatial compression.

We report the results in \cref{tab:comp}. We compare the trade-off between bits per pixel (BPP) and the quality of the compressed data measured in weighted RMSE and weighted MAE.\footnote{Weights are assigned based on varying cell areas as in training.} Here, one can observe that our method shows superior results to spatial compression baselines. Our method achieves comparable results with FFN-T, which is a spatial-temporal compression algorithm that is not directly comparable to ours. In particular, FFN-T takes advantage of sharing parameters across different time steps, while our scheme can easily incorporate the daily growth of the dataset.\footnote{Our scheme can train a small network only for the new data, while FFN-T requires re-training on the updated dataset.} 

\subsection{Cosmic Microwave Background Data}\label{subsec:exp2}
Here, we evaluate our model for the CMB data, which is stored as a HEALPix grid where the distribution of the data is uniform across the grids. For every CMB experiment, we used a resolution of $N_{\text{side}}=2048$, which is the highest resolution that the Planck satellite provides. One should note that we used the highest resolution of the CMB data (i.e., $N_\text{side}$ = 2048) where the data is often downgraded or only utilizes part of the map due to its extensive data size \citep{krachmalnicoff2019convolutional, montefalcone2021inpainting}.  Meanwhile, our method utilizes the full resolution and highlights the scalability and ability to learn intricate signals of the CMB data.

\textbf{Super-resolution. }Similar to the regression task, we can apply super-resolution for CMB data. However, the experiment setting is slightly adjusted for CMB data since the resolution of HEALPix data is defined by $N_{\text{side}}$. Here, the $\times 2$ setting involves training on $N_{\text{side}}=1024$ to predict $N_{\text{side}}=2048$, and the $\times 4$ setting trains the model on $N_{\text{side}}=512$ to predict $N_{\text{side}}=2048$. 

Note that training on high-resolution $N_{\text{side}}=2048$ data is challenging due to its large size, a hurdle our CNR-based approach overcomes, unlike prior works that often use downgraded data such as $N_{\text{side}}=64$~\cite{defferrard2020deepsphere}. As can be seen in the \cref{tab:cosmo} our approach outperforms the other baselines. Despite the highly intricate and noisy pattern of the CMB data, our model significantly outperforms the baselines. This implies that our model can learn complex real-world data as well.

\begin{figure}[t]

    \centering
    \captionof{table}{\textbf{Results of CMB data.} Sup. indicates super-resolution and Reg. indicates regression. The evaluation is based on PSNR and the best metric is highlighted in \textbf{bold}.}
    \label{tab:cosmo}
    \vspace{0.1in}
     \begin{tabular}{lcccc}
        \toprule
         & Sup. ($\times 2$) & Sup. ($\times 4$) &  Reg. \\
        \midrule
        ReLU+P.E. & 21.30 & 22.48 & 24.44\\ 
        SIREN & 21.50 & 23.42 & 28.30\\ 
        SHINR & 21.11 & 22.16 & 22.67 \\ 
        WIRE & 26.01 & 26.51  & 30.09\\ 
        FFN & 23.98 & 24.88  & 28.39\\ 
        \midrule
        \textbf{HNeR-S (ours)} & \textbf{32.65} &\textbf{30.37} &  \textbf{38.72}\\
        \bottomrule
    \end{tabular}
\end{figure}

 We provide experimental results in \cref{tab:cosmo} and \cref{fig:vis_cos}. We observe that our method consistently shows superior PSNR results. Moreover, in \cref{fig:vis_cos}, our model converges into desirable PSNR in the early stages of the training. 

\textbf{Regression. }Similar to the regression in climate data, we predict the unknown signal from the constructed hybrid neural representation. We demonstrate the results in \cref{tab:cosmo}. One can observe that our method highly outperforms compared to the baselines. As depicted in \cref{fig:vis_cos}, both the prediction and error map indicate that our method can predict the target signal within a small error. Moreover, according to \cref{fig:psnr}, our model converges to the promising accuracy by showing a large performance gap compared to the other baselines. 

\section{Conclusion}
In this paper, we presented two new hybrid neural representations tailored for spherical data, utilizing hierarchical spherical feature-grids to generate positional encodings through spherical interpolation algorithms. Our extensive experimental evaluations across multiple datasets and tasks have verified the effectiveness of our method. An interesting avenue for future work will be the application of adaptive grid \citep{martel2021acorn, martinez2022multiresolution} which will lead to a more compact feature grid. Also, there remains an exploration for interpolation methods tailored for spherical grids, i.e., geodesic-aware interpolation.


\clearpage
\newpage
\section*{Broader Impact}
Our framework can advance machine learning for applications like climatology and cosmology. Machine learning for weather and climate forecasting can benefit our society via improved resilience to extreme weather events, agriculture productivity, and water resource management. Furthermore, machine learning for cosmology can accelerate scientific discovery, in particular our understanding of dark matter and dark energy.

\section*{Acknowledgements}
We would like to express our sincere gratitude to Seonghyun Park, Hyosoon Jang, Seongsu Kim, Juwon Hwang, Kiyoung Seong, and Dongyeop Woo for their valuable comments and offering essential experimental tools for our research. Also, we are grateful to Langwen Huang for providing the detailed results of their work, which served as a foundational baseline method for our research. Additionally, our thanks extend to Jihoon Tack and Doyup Lee for their constructive feedback that greatly contributed to the improvement of this paper.

\bibliography{example_paper}
\bibliographystyle{icml2024}

\newpage
\appendix
\onecolumn

\section{Architecture Details}

\textbf{Equirectangular feature-grid.} For the equirectangular feature-grid we adjust the level of the hierarchical map and the base resolution of the feature-grid, by considering the resolution of the training data. For example, comparing the task of $\times$2 and $\times$4 super-resolution, we assign the higher level and base resolution for $\times$2 task. Mainly, we used upscaling factor $\gamma=1.5$. We initialized the parameter for the feature-grid with uniform initialization $\mathcal{U}(-1\mathrm{e}{-4},1\mathrm{e}{-4})$.

\textbf{HEALPix feature-grid.} Similarly to the equirectangular feature-grid, we adjust the feature grid level by the resolution of the training data. The choice of the level is quite intuitive. In the case for super-resolution and regression task, mostly the optimal choice was to choose the level as $\ell_{\text{grid}}=\ell_{\text{data}}-1$. This choice was beneficial since it enables overall parameters to be trained while capturing the fine details. The parameter of the feature-grid is initialized by normal initialization $\mathcal{N}(0,1\mathrm{e}{-4})$. 
\section{Baseline Details}

\textbf{ReLU+P.E.}~\cite{mildenhall2021nerf} is a positional encoding defined as: 
\begin{equation*}
\gamma(x)=\left(\sin \left(2^0 \pi x\right), \cos \left(2^0 \pi x\right), \cdots, \sin \left(2^{L-1} \pi x\right), \cos \left(2^{L-1} \pi x\right)\right)
\end{equation*}
which separately encodes each dimension of the input point $\mathbf{x}\in\mathbb{R}^n$ with sinusoidals in a axis-aligned and logarithmically spaced manner. In this study, we adjust the level of the frequency $L$ with $\lbrace 2, 5, 7\rbrace$. 

\textbf{SIREN}~\cite{sitzmann2020implicit} is a sinusoidal activation $\sin(\omega_0 W\mathbf{x}+b)$ where the frequency of a sinusoidal function is composed by linear weight and scaled by $\omega_0$. Here, $\omega_0$ is set as a hyperparameter and we adjust its range within $\lbrace 10, 20, 30 \rbrace$. 

\textbf{FFN}~\cite{tancik2020fourier} is a random fourier feature positional encoding $\gamma\left(\mathbf{x})=((\cos(2\pi\mathbf{B}\mathbf{x}),\sin(2\pi\mathbf{B}\mathbf{x})\right)^T $ where $\mathbf{x}\in\mathbb{R}^n$ and $\mathbf{B}\in \mathbb{R}^{m\times n}$ is a sampled frequency from $\mathcal{N}(0, \sigma^2)$. In our study, we adjust $\sigma$ in the range of $\lbrace 2, 5, 10\rbrace$.

\textbf{WIRE}~\cite{saragadam2023wire} is a Gabor wavelet based activation $\exp({j\omega_0 \mathbf{x}})\exp({-|s_0 \mathbf{x}|^2})$, which assigns the scale ($\omega_0$) and width ($s_0$) as hyper parameters. In our study, we adjust the both scale and width parameters in the range of $\lbrace 1, 10\rbrace$.

\textbf{SHINR}~\cite{esteves2022generalized} is a spherical harmonics based positional encoding $\gamma(\mathbf{x})=\bigoplus_{l=0}^L\bigoplus_{m=-l}^l Y_m^l(\theta,\psi)$, which takes spherical coordinate $(\theta,\psi)$ and $L$ is a level of the encoding. In our experiment, we adjust the level $L$ in the range of $\lbrace 2, 3, 4\rbrace$.
\section{Experimental Details}\label{appx:experiment}
\subsection{Hyperparameters.}
In this section, we present the spectrum of hyperparameters applied to our approach. Without loss of generality, the hyperparameters selected for weather and climate datasets correspond to those used for the equirectangular feature grid, while those chosen for cosmic microwave background data pertain to the HEALPix feature grid. The common settings across the tasks are 4 hidden layers, 256 hidden dimensions, ReLU \cite{agarap2018deep} activation and AdamW optimizer  \cite{loshchilov2017decoupled} for the MLP.

\begin{table}[!h]
\caption{Hyperparameter table for super-resolution task for geopotential, temperature and CMB data.}\label{tab:hyperparamerter_sr}
\centering
\resizebox{\textwidth}{!}{
    \begin{tabular}{lcccccc}
        \toprule
        \textbf{INR training}  & \textbf{Geo. SR ($\times$ 2)} & \textbf{Geo. SR ($\times$ 4)}&\textbf{Temp. SR ($\times$2)} & \textbf{Temp. ($\times$4)} &  \textbf{CMB ($\times$2)} & \textbf{CMB ($\times$4)} \\
        \midrule
        Level ($L$)  &$\lbrace 8, 6\rbrace$ & $\lbrace 8, 6\rbrace$ &$\lbrace 8, 6\rbrace$ & $\lbrace 8, 6\rbrace$ &$\lbrace 11, 10\rbrace$ & $\lbrace 10, 9\rbrace$ \\
        Parameter dim. ($d$) &$\lbrace 2, 3\rbrace$ & $\lbrace 2, 3\rbrace$ &$\lbrace 2, 3\rbrace$ & $\lbrace 2, 3\rbrace$ &$\lbrace 2, 3\rbrace $ & $\lbrace 2, 3\rbrace$ \\
        Scaling factor ($\gamma$) & $1.5$ & $1.5$ &$1.5$ & $1.5$ &$-$ & $-$  \\
        Base resolution  & $16$ & $16$ &$16$ & $16$ &$-$ & $-$ \\
        \bottomrule
    \end{tabular}%
    }
\end{table}
\begin{table}[!h]
\caption{Hyperparameter table for regression and temporal task for geopotential, temperature and CMB data. \textbf{1 day} stands for daily and \textbf{1 week} stands for weekly interpolation task.}\label{tab:hyperparamerter_reg_temp}
\centering
\resizebox{\textwidth}{!}{
    \begin{tabular}{lccccccc}
        \toprule
        \textbf{INR training} & \textbf{Geo. (Reg.)} & \textbf{Temp. (Reg.)} &  \textbf{CMB (Reg.)} & \textbf{Geo. (1 day)} & \textbf{Temp. (1 day)} & \textbf{Geo. (1 week)} & \textbf{Temp. (1 week)} \\
        \midrule
        Level ($L$) &$\lbrace 8, 6\rbrace$ & $\lbrace 8, 6\rbrace$  & $\lbrace 11, 12\rbrace$ & $\lbrace 7, 8\rbrace$ & $\lbrace 6, 8\rbrace$& $\lbrace 8\rbrace$ & $\lbrace 6, 8\rbrace$\\
        Parameter dim. ($d$)& $\lbrace 2, 3\rbrace$ & $\lbrace 2, 3\rbrace$ & $\lbrace 2, 3\rbrace$  & $\lbrace 6, 8\rbrace$ & $\lbrace 2, 3\rbrace$& $\lbrace 8,9\rbrace$ & $\lbrace 2, 3\rbrace$\\
        Scaling factor ($\gamma$) & $1.5$ & $1.5$ & $-$ & $1.5$& $1.5$& $\lbrace 1.1, 1.2\rbrace$ & $1.5$ \\
        Base resolution  & $16$ & $16$   & $-$ & $16$ & $4$& $4$ & $16$\\ 
        \bottomrule
    \end{tabular}%
    }
\end{table}


\subsection{Training Details.} We train our model with weighted RMSE \cite{huang2023compressing} and used AdamW optimizer \cite{loshchilov2017decoupled} with  learning rate of 1e-5.  For the compression task, we trained our model with~. We utilize the C3 \citep{kim2023c3} framework for compression task which also utilizes the learnable positional encoding to encode the data. Specifically, we applied our equirectangular grid feature instead of the original learnable positional grid which is targeted for a single or a series of 2D image data.

\subsection{Task Details.} 
\textbf{Super-resolution.} The study applies two settings for enhancing the resolution of CMB, weather, and climate data: $\times2$ and $\times4$. In the $\times2$ setting, the model is trained using $361 \times 720$ training points, each representing a 0.50$^\circ$ degree. For the $\times4$ setting, training involves $181\times 360$ points, with each point representing a 1.00$^\circ$ degree. The model's performance is assessed on its ability to predict data at a finer resolution of 0.25$^\circ$, across a total of $721\times 1440$ points. In the context of CMB super-resolution, the $\times2$ setting involves training on data downsampled to $N_\text{side}=1024$, yielding approximately 10M training points at a resolution of about $N\mu$min. For the $\times4$ setting, the training data is further downsampled to $N_\text{side}=512$, with around 30K points. The prediction task targets data with $N_\text{side}=2048$, encompassing about 50M points at the full resolution provided by the Planck satellite.

\textbf{Regression.} The regression task involves dividing data points into training and testing sets in an 8:2 ratio. Specifically, for weather and climate data, the split is conducted on geopotential and temperature datasets that total $721\times 1440$ points, resulting in approximately 83K training points and 20K test points. For CMB data, the split includes 50M points, allocating 40M for training and 10M for testing.

\textbf{Temporal interpolation.} This task employs two temporal resolutions: daily and weekly, with a total of 30 data snapshots sampled. Here, snapshots indicate the data that represents the captured moment. The spacing between snapshots corresponds to the chosen temporal resolution. Training data is constructed by selecting the first snapshot, skipping the next, and then selecting the third snapshot. Conversely, test data is compiled by skipping the first snapshot, selecting the second, and continuing this pattern. This method ensures a consistent and desired time step between training and test datasets.

\textbf{Compression. }The two settings of the compression task are divided by the bits per pixel (BPP). The BPP is computed by dividing the model size by the number of pixels. On one hand, for the spatiotemporal compression task, the number of pixels is the multiplication of the number of time indices, pressure levels, and the number of latitude and longitude levels. In our dataset, they are 366, 11, 361, and 720, respectively, following \citep{huang2023compressing}. On the other hand, for the spatial compression task, the number of pixels is the multiplication of the number of latitude and longitude levels.

\newpage

\end{document}